\documentclass[11pt,letterpaper]{article}
\usepackage[margin=1in]{geometry}\usepackage{lmodern}
\usepackage[T1]{fontenc}
\usepackage[utf8]{inputenc}
\usepackage{cite}
\usepackage{amsmath,amssymb,amsfonts}
\usepackage{graphicx}
\usepackage{booktabs,array,tabularx,makecell,multirow}
\usepackage{longtable}
\usepackage{supertabular}
\usepackage{needspace}
\usepackage{capt-of}
\usepackage{stfloats}
\usepackage{placeins}
\usepackage{textcomp}
\usepackage{url}
\usepackage[hidelinks]{hyperref}
\usepackage{microtype}
\newcolumntype{L}[1]{>{\raggedright\arraybackslash}p{#1}}
\newcolumntype{Y}{>{\raggedright\arraybackslash}X}

\providecommand{\hl}[1]{#1}
\newcommand{\tablefont}{\small}
\newcommand{\tablenote}[1]{\par\vspace{4pt}\raggedright\footnotesize #1}
\hypersetup{pdftitle={COMBIO: Automated Discovery of a Compact Composite Biomarker from Routine Laboratory Panels},pdfauthor={Tomer D. Meirman; Lior S. Rokach; Bracha Shapira; Seffi Cohen; Noa Dagan}}

\begin{document}
\title{COMBIO: Automated Discovery of a Compact Composite Biomarker from Routine Laboratory Panels}
\author{Tomer~D.~Meirman, Lior~S.~Rokach, Bracha~Shapira, Seffi~Cohen, and Noa~Dagan\thanks{T.~D.~Meirman, L.~S.~Rokach, B.~Shapira, S.~Cohen, and N.~Dagan are with the Stein Faculty of Computer and Information Sciences, Ben-Gurion University of the Negev, Beer-Sheva, Israel. N.~Dagan is also with the Clalit Research Institute, Clalit Health Services, Ramat Gan, Israel. L.~S.~Rokach: \texttt{liorrk@post.bgu.ac.il}. B.~Shapira: \texttt{bshapira@bgu.ac.il}. S.~Cohen: \texttt{seffic@bgu.ac.il}. N.~Dagan: \texttt{noada@clalit.org.il}. Corresponding author: T.~D.~Meirman (e-mail: \texttt{meirmant@post.bgu.ac.il}).}}
\date{}\maketitle
\begin{abstract}Machine-learning models based on routine laboratory panels can predict outcomes but can be difficult to interpret and operationalize. We present COMBIO, an automated pipeline that distils such a model into a compact closed-form biomarker. The composite is a product of laboratory measurements raised to integer powers from minus two to two. The sign of each exponent indicates whether higher values of the corresponding measurement increase or decrease the composite. Shapley additive explanation values from the gradient-boosted reference model rank five candidate measurements and guide eligible exponent signs. An exhaustive search over small integer exponents selects the formula on an inner validation set, before evaluation on the reserved test set. We evaluated COMBIO across six tasks: three National Health and Nutrition Examination Survey outcomes, liver-disease classification in the Indian Liver Patient Dataset, and Medical Information Mart for Intensive Care IV in-hospital mortality and sepsis case-finding, with external evaluation in the eICU Collaborative Research Database for the two hospital tasks. Across the six test sets, COMBIO's area under the receiver operating characteristic curve differed by at most 0.03 from logistic regression using the five candidate laboratory measurements. The hospital mortality composite had an area under the curve of 0.698 in the external database. A version restricted to widely available measurements in the source hospital database achieved an area under the curve of 0.786 in a cohort covering 92\% of admissions. The multiplicative representation provides compactness rather than additional accuracy, and local validation and recalibration are required before clinical use.\end{abstract}
\section{Introduction}\label{introduction}

\subsection{Background}\label{background}

Machine-learning risk models based on routine laboratory panels often achieve strong discrimination, but their complexity can hinder interpretation and bedside implementation. Compact, manually constructed measures developed for specific clinical purposes include the neutrophil-to-lymphocyte ratio \cite{r1}, the fibrosis-4 index (FIB-4) \cite{r2}, and the triglyceride-glucose index \cite{r3}. These measures are familiar and readily calculated. However, the broader routine laboratory panel may contain predictive information that is not captured by any single marker or existing index. A method is therefore needed to translate the signal learned by a complex model into a comparably compact, transparent, and directly calculable biomarker. In this work, we present COMBIO, an automated method designed to perform this translation.

\subsection{Prior Work}\label{prior-work}

\textbf{Hand-crafted clinical indices.} Several compact clinical indices combine a small number of routine laboratory measurements into a single score or ratio that is easy to calculate and communicate. Each index was developed or evaluated for a specific purpose. FIB-4 \cite{r2} and the aspartate aminotransferase-to-platelet ratio index (APRI) \cite{r4} estimate liver fibrosis. The neutrophil-to-lymphocyte ratio (NLR) \cite{r1} indicates systemic inflammation and physiological stress. The Onodera prognostic nutritional index (PNI) \cite{r5} combines serum albumin and peripheral lymphocyte count to estimate nutritional status and operative risk. The triglyceride-glucose index (TyG) \cite{r3} is a surrogate for insulin resistance. The triglyceride-to-high-density-lipoprotein cholesterol ratio (TG/HDL) \cite{r6} has been studied in relation to myocardial infarction risk. Red cell distribution width (RDW) \cite{r7} is associated with mortality. We compare COMBIO with selected indices both in their published forms and after fitting logistic models to their component variables within each cohort.

\textbf{Learned interpretable scoring systems.} Several methods learn compact prediction rules from data. Supersparse linear integer models and RiskSLIM \cite{r8,r9}, FasterRisk \cite{r10}, and AutoScore \cite{r11} produce sparse additive point scores with integer weights. SHAPoint \cite{r12} uses Shapley additive explanations (SHAP) values \cite{r22} to select and rank variables and to translate model contributions into an additive integer point score. Optimal sparse decision trees \cite{r13} provide tree-based representations, whereas generalized additive models with pairwise interactions (GA\textsuperscript{2}M) \cite{r14}, including explainable boosting machines \cite{r15}, provide additive shape-function representations. COMBIO likewise uses model-based feature ranking, but its output differs: it is a single multiplicative formula applied directly to laboratory measurements rather than an additive point score. The multiplicative representation is therefore intended to provide a compact closed-form biomarker, not additional predictive accuracy.

\textbf{Symbolic regression for clinical scores.} Symbolic regression searches a user-defined space of mathematical expressions composed of variables and operators to identify equations that predict an outcome \cite{r16,r17,r18}. Recent studies have applied symbolic regression to interpretable clinical prediction models (the feature engineering automation tool, FEAT \cite{r19}), automated clinical score generation (SymScore \cite{r20}), and clinical decision scores that combine equations and rules \cite{r21}. COMBIO restricts this broad search space to a single monomial, defined as one product term with small integer exponents, constructed from laboratory variables ranked by the reference model. This restricted hypothesis space can be exhaustively searched and yields a compact, structurally consistent expression rather than a less constrained equation. In this study, we additionally evaluate calibration, decision-curve performance, and external transportability across independent cohorts and hospitals.

\textbf{Explainability and distillation.} Post-hoc methods such as SHAP \cite{r22} assign feature-attribution values to predictions from a fitted model but do not replace the model used for prediction. Model distillation transfers predictive information from a complex teacher model to a simpler student model \cite{r23,r24}. The student nevertheless remains a predictive model. COMBIO uses the reference model to guide discovery of a closed-form biomarker that can itself be used for risk stratification. The resulting formula remains a predictive model. Tree SHAP guides variable ranking and constrains exponent directions in COMBIO. The formula search uses outcome labels rather than teacher predictions. Reporting is organized using the Transparent Reporting of a multivariable prediction model for Individual Prognosis Or Diagnosis guidance for artificial intelligence (TRIPOD+AI) \cite{r25} (Supplementary Note 1).

\subsection{Study Objectives}\label{study-objectives}

We present COMBIO (composite monomial biomarker), a reproducible method that distils a risk model based on routine laboratory measurements into a single closed-form composite defined by a product of a few measurements raised to small integer powers. We evaluate the method using a transparently reported six-task benchmark. The study has four objectives. First, we define an automated procedure for selecting the component measurements and integer exponents, using Tree SHAP magnitude and directional trends in the inner validation subset. Second, we compare COMBIO with the strongest observed single laboratory measurement among five SHAP-selected candidates, published clinical indices, refitted models using index components, logistic regression using the five candidate measurements, and a full XGBoost model using broader inputs. Third, we assess external and cross-hospital transportability, calibration, and decision-curve performance. Fourth, we provide parameterized code for independent replication. COMBIO is a constrained special case of logistic regression applied to log-transformed versions of the candidate laboratory variables. Accordingly, the multiplicative form is intended as a compact and directly calculable representation, not as a means of exceeding the accuracy of a full multivariable model. The study therefore evaluates the performance, transportability, and limitations of this representation rather than proposing a change to clinical practice.

\section{Methods}\label{methods}

\subsection{Data Sources and Cohorts}\label{data-sources-and-cohorts}

We evaluated COMBIO in six tasks spanning screening, diagnosis, and prognosis (Table~\ref{tab1}). Three tasks used adults aged 18 years or older in the National Health and Nutrition Examination Survey (NHANES) \cite{r26}. For iron overload, we used NHANES 2017--18 and defined an analysis phenotype requiring transferrin saturation \ensuremath{\geq}45\% and ferritin \textgreater300 ng/mL in men or \textgreater200 ng/mL in women.

For five-year all-cause mortality, we used NHANES 2009--10 and its public-use linked mortality file, with NHANES 2005--06 for external evaluation. Deaths within 60 months were cases, follow-up beyond 60 months served as controls, and records with insufficient follow-up were excluded.

For diabetes, we used NHANES 2017--18 for primary derivation and 2015--16 for external evaluation, with a separate pooled-cycle robustness analysis. The diabetes outcome was a single glycated hemoglobin (HbA1c) measurement \ensuremath{\geq}6.5\% among participants reporting no prior diagnosis \cite{r27}. Because this label relied on one HbA1c measurement, it was an analysis-defined phenotype rather than a confirmed clinical diagnosis.

The fourth task used the Indian Liver Patient Dataset (ILPD) from the University of California, Irvine (UCI) Machine Learning Repository \cite{r28}, a case-enriched diagnostic dataset in which the prediction task was to distinguish patients with liver disease from those without liver disease using routine laboratory measurements.

Finally, the two hospital tasks used the Medical Information Mart for Intensive Care IV (MIMIC-IV) \cite{r29}: in-hospital mortality and sepsis case-finding, defined as identifying hospitalisations carrying an International Classification of Diseases (ICD) sepsis diagnosis from recorded laboratory measurements. Both MIMIC-IV tasks were externally evaluated in the eICU Collaborative Research Database, which contains data from more than 200 US hospitals \cite{r30}. The prediction inputs were routine laboratory measurements recorded in the source datasets.

\begin{table}[!t]
\caption{Source cohorts and outcome counts}
\label{tab1}
\centering\tablefont
\setlength{\tabcolsep}{3pt}
\begin{tabular}{@{}L{\dimexpr 0.3300\linewidth-1.9800\tabcolsep\relax} L{\dimexpr 0.1800\linewidth-1.0800\tabcolsep\relax} L{\dimexpr 0.2200\linewidth-1.3200\tabcolsep\relax} L{\dimexpr 0.2700\linewidth-1.6200\tabcolsep\relax}@{}}
\toprule
\textbf{Task and source} & \textbf{Cohort N} & \textbf{Cohort events (\%)} & Common test N /\allowbreak{} events \\
\midrule
Iron overload. NHANES 2017--18 & 5,122 & 113 (2.2\%) & 1,532 /\allowbreak{} 34 \\
\addlinespace[2pt]
5-year mortality. NHANES 2009--10 & 6,346 & 363 (5.7\%) & 1,772 /\allowbreak{} 94 \\
\addlinespace[2pt]
Diabetes. NHANES 2017--18 & 4,290 & 151 (3.5\%) & 1,253 /\allowbreak{} 43 \\
\addlinespace[2pt]
Liver disease. ILPD & 583 & 416 (71.4\%) & 175 /\allowbreak{} 125 \\
\addlinespace[2pt]
In-hospital mortality. MIMIC-IV & 434,542 & 10,596 (2.4\%) & 28,988 /\allowbreak{} 1,891 \\
\addlinespace[2pt]
Sepsis case-finding. MIMIC-IV & 434,542 & 22,109 (5.1\%) & 28,858 /\allowbreak{} 3,898 \\
\bottomrule
\end{tabular}
\tablenote{Cohort events refer to the full outcome-labelled cohorts. The test column reports the common-case test subset for each task. Analysis-cohort flow and age/sex characteristics are in Supplementary Tables~\ref{stab1} and \ref{stab2}. NHANES, National Health and Nutrition Examination Survey; ILPD, Indian Liver Patient Dataset; MIMIC-IV, Medical Information Mart for Intensive Care IV.}
\end{table}

\subsection{Composite Biomarker Definition and Discovery}\label{composite-biomarker-definition-and-discovery}

\begin{figure*}[!t]
\centering
\includegraphics[width=\textwidth]{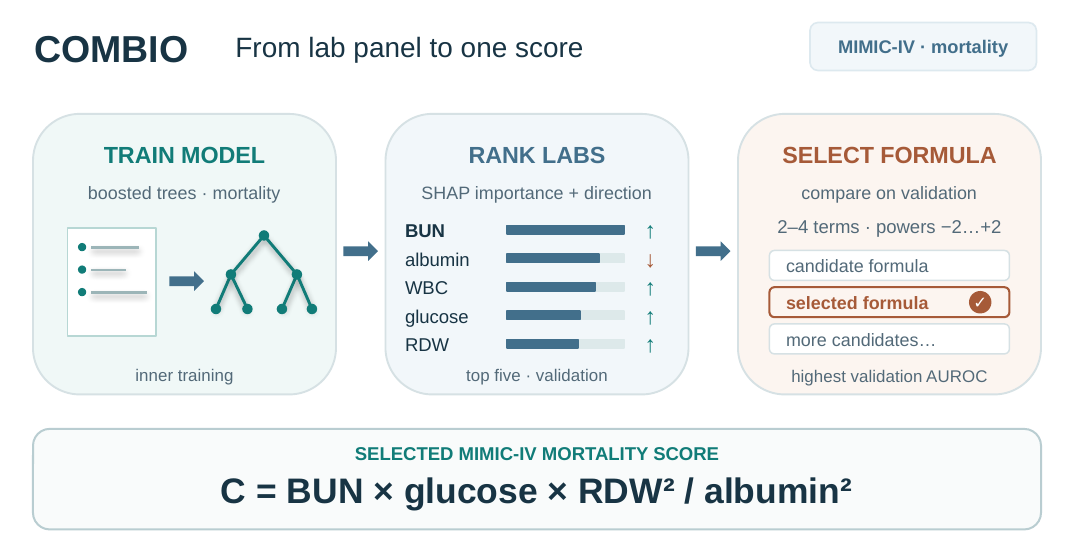}
\caption{COMBIO formula discovery, illustrated by MIMIC-IV in-hospital mortality. A model fitted in inner training supplies mean absolute Tree SHAP values and directional trends in inner validation. These guide measurement ranking and eligible integer exponents for formulas with two to four terms. Validation AUROC selects the formula, with risk orientation set using inner training labels. Test data are reserved before fitting and used after selection. BUN, blood urea nitrogen; WBC, white blood cell count; RDW, red cell distribution width; SHAP, Shapley additive explanations. Directional trends describe model associations rather than causal effects. The selected example is \(C=\mathrm{BUN}\times\mathrm{glucose}\times\mathrm{RDW}^{2}/\mathrm{albumin}^{2}\); WBC receives exponent zero.}
\label{fig1}
\end{figure*}

COMBIO defines a composite \(C = \prod_{i}^{}x_{i}^{a_{i}}\), where \(x_{i}\) denotes the ith laboratory measurement and \(a_{i}\) is its small integer exponent (\(a_{i}\in\mathbb{Z}\)). Taking logarithms gives \(\log C = \sum_{i}^{}a_{i}\log x_{i}\). Thus, the composite is equivalent to a linear score in log-transformed laboratory measurements, with each exponent determining the magnitude and direction of that measurement\textquotesingle s contribution. Discovery proceeds in three stages (Fig.~\ref{fig1}): training the reference model, ranking measurements and assessing their directions, and selecting the formula. Before fitting, we reserve 30\% of each source cohort as a test set. The remaining source observations are split 70:30 into inner training and validation subsets. Available external cohorts remain separate and do not inform formula selection. Both splits keep all MIMIC-IV admissions from the same patient together.

First, a gradient-boosted reference model is fitted to the eligible routine laboratory panel in inner training. Second, we calculate Tree SHAP values \cite{r22} for the positive-class log odds on inner validation. We rank eligible measurements by mean absolute SHAP value and retain the top five (Supplementary Table~\ref{stab3}). Measurements used to define the outcome are excluded. Glucose is also excluded for diabetes to assess a glucose-free alternative. For each selected measurement, we assess its directional association by relating its values to its SHAP contributions. A positive or negative direction is assigned only when the Spearman rank correlation \cite{r31} between its measured value and SHAP contribution and the difference between upper- and lower-quartile mean SHAP values agree in sign, and the mean SHAP values across five ranked groups do not reverse trend. That direction restricts the sign of the final risk-oriented exponent. When these checks disagree, both signs remain eligible. Third, we enumerate candidate formulas by assigning each selected measurement an exponent \(a_{i} \in \{ - 2, - 1,0,1,2\}\), retaining formulas with two to four nonzero terms and eligible signs. Inner training labels set each candidate formula\textquotesingle s risk orientation. We evaluate these formulas on inner validation and select the one with the highest area under the receiver operating characteristic curve (AUROC) rounded to four decimal places. If rounded AUROCs tie, we prefer fewer nonzero terms. After discovery, the selected formula and orientation are locked before use on the test set and independent external cohorts. We call the resulting score COMBIO.

\subsection{Comparators}\label{comparators}

We evaluated five comparator classes, ordered from least to most flexible: (i) the strongest observed single laboratory measurement among the five SHAP-selected candidates; (ii) relevant published clinical indices in their original forms; (iii) logistic refits on each index\textquotesingle s log-transformed components, fitted in training; (iv) logistic regression on all five SHAP-selected log-transformed measurements with unconstrained coefficients; and (v) a full XGBoost model \cite{r32} using the task-specific laboratory panel and available demographics, fitted on combined training and validation data. The strongest observed candidate is reported descriptively because its identity was determined by test AUROC.

Additional inputs were age, sex, race, body mass index and waist circumference in NHANES, and age and sex in MIMIC-IV. ILPD used laboratory measurements only. Full XGBoost was evaluated for all six tasks. MIMIC-IV age is the recorded de-identified anchor\_age, not age reconstructed for each admission \cite{r29}. The reference model supplies Tree SHAP attribution for candidate ranking, whereas the full comparator estimates performance with broader inputs and greater flexibility. COMBIO retains two to four of the five candidates, so the five-variable logistic benchmark can use candidates omitted from the formula.

As a secondary comparison for iron overload, ILPD liver disease and MIMIC-IV mortality, we fitted SHAPoint \cite{r12}, FasterRisk \cite{r10}, an explainable boosting machine (EBM) \cite{r15} and gplearn symbolic regression \cite{r17} using the source training partition (combined inner training and validation). SHAPoint selected five variables from the full laboratory panel. FasterRisk used training-quartile binary indicators for that panel and up to six score terms. The additive boosting model used the full panel without interactions. Symbolic regression was fitted both to the full panel and separately to COMBIO\textquotesingle s retained measurements after log transformation. MIMIC-IV training was capped at 30,000 observations for SHAPoint, 60,000 for FasterRisk and additive boosting, and 20,000 for symbolic regression, sampled with seed 42. Where needed, missing values were replaced with fitting-subset medians. SHAPoint handled missing values natively. Any score orientation was fixed in training. All learned scorers and COMBIO were evaluated on the same task-specific complete-case test subset, with 95\% AUROC intervals from 1,000 participant or admission bootstrap resamples.

The clinical-index candidates include the De Ritis aspartate aminotransferase (AST)/alanine aminotransferase (ALT) ratio \cite{r33}, C-reactive protein (CRP)/albumin \cite{r34}, blood urea nitrogen (BUN)/albumin \cite{r35}, NLR \cite{r1,r36}, and TG/HDL \cite{r6,r37}. The evaluated indices, their component measurements, and their original clinical uses are defined in Supplementary Table~\ref{stab4}. PNI is calculated as 10 \ensuremath{\times} albumin (g/dL) + 0.005 \ensuremath{\times} absolute lymphocyte count (cells/\ensuremath{\mu}L) \cite{r5}. Lower values generally indicate poorer nutritional and immune status. PNI has been evaluated for mortality in community and critically ill cohorts \cite{r38,r39}. The De Ritis ratio is AST/ALT, originally described in viral hepatitis \cite{r33}. APRI was developed for liver-fibrosis assessment \cite{r4}. FIB-4 was developed to assess liver fibrosis \cite{r2}. Its reported association with mortality \cite{r40} motivated its inclusion as an exploratory comparator. The risk orientation of published indices is fixed using training data. All five single-measurement candidates and each clinical index are evaluated.

TG/HDL is evaluated only in participants with HDL available. COMBIO is reported on that same HDL-complete subset.

\subsection{Evaluation and Statistical Analysis}\label{evaluation-and-statistical-analysis}

Primary comparisons within each task use a common complete-case subset of its reserved test partition, requiring all five SHAP-selected candidates and the components of the evaluated clinical indices. Secondary calibration and highest-risk 5\% and 10\% analyses require only the final formula measurements. External COMBIO--index comparisons require the measurements used by both scores in the independent cohort. Pooled-cycle diabetes, high-coverage MIMIC-IV, and reverse eICU derivations use their specified source cohorts. Additional index-specific exclusions and sample sizes are reported for scores that remain undefined.

Laboratory values are log-transformed after clipping at 0.001 in their recorded units. COMBIO is scored on complete cases. Logistic comparators fit median imputation and standardization within training, then use the same common complete-case test subset as COMBIO. MIMIC-IV splits keep all admissions from a patient together.

As a separate sensitivity comparison, rounded logistic regression fits the task\textquotesingle s laboratory panel in inner training, rounds the largest two to four log-scale coefficients to exponents from \ensuremath{-}2 to +2 and chooses the term count on inner validation. Its score direction is set in training. We compare it with the selected COMBIO score on shared reserved test observations. Exploratory MIMIC-IV subgroup summaries use formula-complete reserved test admissions grouped by recorded sex, age and race.

An exploratory sensitivity analysis evaluated the effect of allowing age alongside laboratory measurements in the five-year-mortality candidate panel. The original inner training, validation, and test partitions and the same SHAP ranking and exponent-search procedure were retained. The age-eligible formula, the primary laboratory-only formula, and FIB-4 were evaluated on a shared test subset. This analysis does not replace the primary laboratory-only benchmark.

We assess discrimination using AUROC and area under the precision--recall curve (AUPR). Calibration is assessed using slope, intercept and Brier score. Calibration plots show observed decile event rates and the fitted diagnostic relation \(\operatorname{expit}(\alpha+\beta\operatorname{logit}(p))\), where \(\alpha\) and \(\beta\) are the estimated calibration intercept and slope, and \(p\) is the source-fitted predicted risk \cite{r41}. The diagnostic fit does not update the reported predictions. Clinical utility is assessed using decision-curve net benefit \cite{r42}. We also report sensitivity, specificity, positive predictive value, negative predictive value and fold enrichment when the highest-risk 5\% and 10\% are flagged. AUROC comparisons use bootstrap confidence intervals from resampling the same evaluation observations for both scores \cite{r43}. MIMIC confidence intervals resample admissions, which may understate uncertainty when patients have repeated admissions.

Cross-hospital analyses include eICU hospitals with at least 100 stays, 25 outcome events and 25 non-events. Hospital-specific AUROC estimates are pooled on the AUROC scale using a DerSimonian--Laird random-effects analysis \cite{r44}. Heterogeneity is described using I\textsuperscript{2} \cite{r45} and between-hospital standard deviation \ensuremath{\tau}.

\subsection{Ethical Considerations}\label{ethical-considerations}

This study used de-identified public datasets (NHANES and the UCI Indian Liver Patient Dataset) and credentialed de-identified research databases (MIMIC-IV and eICU) under their respective data-use agreements \cite{r26,r28,r29,r30}. No new participant data were collected, and no identifiable information was accessed. Accordingly, no additional institutional review board review or informed consent was required beyond the approvals governing the source databases.

\section{Results}\label{results}

\begin{figure*}[!t]
\centering
\includegraphics[width=\textwidth]{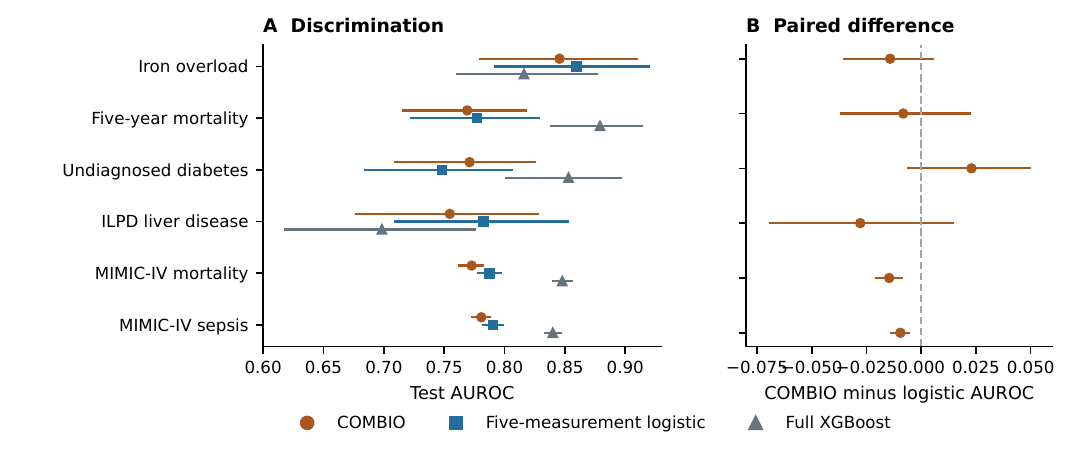}
\caption{Test AUROC for COMBIO, five-candidate logistic regression and full XGBoost. Panel A lines are 95\% percentile bootstrap intervals for every method; panel B shows the paired COMBIO-minus-logistic AUROC difference intervals. Resampling was by participant or admission, as stated in Methods. AUROC, area under the receiver operating characteristic curve.}
\label{fig2}
\end{figure*}

Across six tasks, COMBIO selected formulas containing two to four laboratory measurements. Table~\ref{tab2} and Fig.~\ref{fig2} present the selected formulas and their test discrimination alongside five-variable logistic regression and full XGBoost. COMBIO AUROCs ranged from 0.755 to 0.846 and differed from five-variable logistic regression by no more than 0.03. Table~\ref{tab3} presents comparisons with individual measurements and clinical indices.

The MIMIC-IV in-hospital mortality task illustrates the selected score: BUN \ensuremath{\times} glucose \ensuremath{\times} RDW\textsuperscript{2} / albumin\textsuperscript{2} (Fig.~\ref{fig1}). WBC was one of the five SHAP-ranked candidates, but validation assigned it exponent zero. The formula was fixed before evaluation on the reserved test set.

\begin{table*}[!t]
\caption{\hl{Selected} COMBIO formulas and test discrimination}
\label{tab2}
\centering\tablefont
\setlength{\tabcolsep}{3pt}
\begin{tabular}{@{}L{\dimexpr 0.1300\linewidth-1.3000\tabcolsep\relax} L{\dimexpr 0.2500\linewidth-2.5000\tabcolsep\relax} L{\dimexpr 0.1200\linewidth-1.2000\tabcolsep\relax} L{\dimexpr 0.1700\linewidth-1.7000\tabcolsep\relax} L{\dimexpr 0.1700\linewidth-1.7000\tabcolsep\relax} L{\dimexpr 0.1600\linewidth-1.6000\tabcolsep\relax}@{}}
\toprule
\textbf{Task} & \textbf{COMBIO formula} & \textbf{Test n /\allowbreak{} events} & \textbf{COMBIO AUROC (95\% CI)} & \textbf{Five-laboratory logistic} & \textbf{Full XGBoost} \\
\midrule
Iron overload & GGT \ensuremath{\times} bilirubin \ensuremath{\times} MCV\textsuperscript{2} & 1,532 /\allowbreak{} 34 & 0.846 (0.779--0.910) & 0.860 (0.792--0.919) & 0.816 (0.761--0.876) \\
\addlinespace[2pt]
Five-year mortality & BUN \ensuremath{\times} RDW\textsuperscript{2} \ensuremath{\times} glucose /\allowbreak{} platelets & 1,772 /\allowbreak{} 94 & 0.769 (0.716--0.818) & 0.777 (0.723--0.828) & 0.879 (0.839--0.913) \\
\addlinespace[2pt]
Undiagnosed diabetes & TG\textsuperscript{2} \ensuremath{\times} GGT \ensuremath{\times} hsCRP /\allowbreak{} albumin\textsuperscript{2} & 1,253 /\allowbreak{} 43 & 0.771 (0.710--0.825) & 0.748 (0.685--0.806) & 0.853 (0.802--0.896) \\
\addlinespace[2pt]
ILPD liver disease & alkaline phosphatase\textsuperscript{2} \ensuremath{\times} total bilirubin\textsuperscript{2} \ensuremath{\times} ALT\textsuperscript{2} /\allowbreak{} direct bilirubin & 175 /\allowbreak{} 125 & 0.755 (0.677--0.827) & 0.782 (0.710--0.852) & 0.698 (0.618--0.775) \\
\addlinespace[2pt]
MIMIC-IV mortality & BUN \ensuremath{\times} glucose \ensuremath{\times} RDW\textsuperscript{2} /\allowbreak{} albumin\textsuperscript{2} & 28,988 /\allowbreak{} 1,891 & 0.773 (0.762--0.782) & 0.787 (0.778--0.797) & 0.848 (0.840--0.856) \\
\addlinespace[2pt]
MIMIC-IV sepsis & 1 /\allowbreak{} (lymphocyte \% \ensuremath{\times} albumin\textsuperscript{2} \ensuremath{\times} haemoglobin) & 28,858 /\allowbreak{} 3,898 & 0.781 (0.773--0.788) & 0.790 (0.782--0.798) & 0.840 (0.834--0.846) \\
\bottomrule
\end{tabular}
\tablenote{The formulas were set before test evaluation. AUROC is shown with percentile 95\% bootstrap intervals from 1,000 resamples for every method. Full XGBoost uses the broader task-specific input panel. AUROC, area under the receiver operating characteristic curve; CI, confidence interval; GGT, gamma-glutamyl transferase; MCV, mean corpuscular volume; BUN, blood urea nitrogen; RDW, red cell distribution width; TG, triglycerides; hsCRP, high-sensitivity C-reactive protein; ALT, alanine aminotransferase.}
\end{table*}

\subsection{Primary Discrimination}\label{primary-discrimination}

COMBIO\textquotesingle s relative discrimination was first compared with individual laboratory measurements. COMBIO had a higher AUROC point estimate than the strongest observed member of its five-candidate set in every task (Table~\ref{tab3}). The identity of that member was determined from test AUROC, so this is a descriptive summary rather than a prospectively selected comparator. The difference was clearest in the larger MIMIC-IV mortality and sepsis cohorts. Their bootstrap intervals for COMBIO minus the strongest single candidate were 0.025--0.048 and 0.015--0.029, indicating a significant improvement. Smaller-cohort estimates were less precise, and comparisons were not adjusted for multiplicity.

Five-candidate logistic regression had a higher AUROC than COMBIO in five tasks. COMBIO had a higher point estimate in diabetes (0.771 versus 0.748), where the bootstrap difference interval included zero. Logistic regression was higher in both MIMIC-IV tasks, with COMBIO-minus-logistic difference intervals of \ensuremath{-}0.020 to \ensuremath{-}0.009 for mortality and \ensuremath{-}0.014 to \ensuremath{-}0.006 for sepsis.

Full XGBoost had higher AUROC than COMBIO in four tasks, with its largest point-estimate advantage in five-year mortality (0.879 versus 0.769). Its estimates were lower for iron overload and ILPD (Table~\ref{tab2}). These comparisons do not isolate the monomial constraint because XGBoost used broader inputs.

COMBIO\textquotesingle s relative discrimination was then compared with published clinical indices. Published and refitted indices are separate comparator classes (Table~\ref{tab3}). COMBIO exceeded the three published iron-overload indices. The exploratory FIB-4 mortality comparison is reported in Supplementary Table~\ref{stab5}. In ILPD, published bilirubin/albumin slightly exceeded COMBIO (0.774 versus 0.755), while De Ritis was lower. COMBIO exceeded published BUN/albumin in MIMIC-IV mortality (0.773 versus 0.746) and published NLR in sepsis (0.781 versus 0.709). In the separate HDL-complete diabetes comparison, COMBIO was 0.771 versus 0.675 for TG/HDL.

COMBIO\textquotesingle s relative discrimination compared with refitted clinical-index models. Refitting improved several index point estimates but did not produce a uniform ordering. COMBIO exceeded all three refitted iron indices, both ILPD indices and refitted NLR in sepsis. Refitted BUN/albumin slightly exceeded COMBIO for MIMIC-IV mortality (0.779 versus 0.773). On the HDL-complete diabetes subset, COMBIO exceeded refitted TG/HDL (0.771 versus 0.675). These are task-specific estimates and do not establish superiority over every clinical index.

\FloatBarrier\begingroup\tablefont\setlength{\tabcolsep}{3pt}
\begin{longtable}{@{}L{\dimexpr 0.1700\linewidth-1.3600\tabcolsep\relax} L{\dimexpr 0.2500\linewidth-2.0000\tabcolsep\relax} L{\dimexpr 0.1300\linewidth-1.0400\tabcolsep\relax} L{\dimexpr 0.2250\linewidth-1.8000\tabcolsep\relax} L{\dimexpr 0.2250\linewidth-1.8000\tabcolsep\relax}@{}}
\caption{\hl{Comparisons with} single laboratory measurements and clinical indices}\label{tab3}\\
\toprule
\textbf{Task} & \textbf{Candidate or index} & \textbf{Eligible n} & \textbf{Measured/\allowbreak{}published AUROC (95\% CI)} & \textbf{Refitted AUROC (95\% CI)} \\
\midrule
\endfirsthead
\multicolumn{5}{l}{\textit{Table 3 (continued)}}\\
\toprule
\textbf{Task} & \textbf{Candidate or index} & \textbf{Eligible n} & \textbf{Measured/\allowbreak{}published AUROC (95\% CI)} & \textbf{Refitted AUROC (95\% CI)} \\
\midrule
\endhead
\midrule\multicolumn{5}{r}{\textit{Continued on next page}}\\
\endfoot
\bottomrule\endlastfoot
Iron overload & COMBIO & 1,532 & 0.846 (0.779--0.910) & --- \\
 & Best single: bilirubin & 1,532 & 0.781 (0.710--0.852) & --- \\
 & FIB-4 & 1,532 & 0.708 (0.604--0.801) & 0.786 (0.700--0.857) \\
 & APRI & 1,532 & 0.767 (0.669--0.850) & 0.769 (0.671--0.855) \\
\addlinespace[2pt]
Five-year mortality & COMBIO & 1,772 & 0.769 (0.716--0.818) & --- \\
 & Best single: RDW & 1,772 & 0.741 (0.689--0.787) & --- \\
 & PNI & 1,772 & 0.720 (0.658--0.775) & 0.762 (0.707--0.812) \\
\addlinespace[2pt]
Undiagnosed diabetes & COMBIO & 1,253 & 0.771 (0.710--0.825) & --- \\
 & Best single: hsCRP & 1,253 & 0.725 (0.652--0.786) & --- \\
 & COMBIO (HDL-complete) & 1,252 & 0.771 (0.711--0.831) & --- \\
 & TG/\allowbreak{}HDL (HDL-complete) & 1,252 & 0.675 (0.613--0.738) & 0.675 (0.612--0.739) \\
\addlinespace[2pt]
ILPD liver disease & COMBIO & 175 & 0.755 (0.677--0.827) & --- \\
 & Best single: total bilirubin & 175 & 0.735 (0.660--0.807) & --- \\
 & Bilirubin/\allowbreak{}albumin ratio & 175 & 0.774 (0.703--0.841) & 0.741 (0.667--0.813) \\
 & De Ritis ratio & 175 & 0.657 (0.569--0.741) & 0.739 (0.657--0.816) \\
\addlinespace[2pt]
MIMIC-IV mortality & COMBIO & 28,988 & 0.773 (0.762--0.782) & --- \\
 & Best single: albumin & 28,988 & 0.737 (0.726--0.747) & --- \\
 & BUN/\allowbreak{}albumin ratio & 28,988 & 0.746 (0.735--0.757) & 0.779 (0.769--0.789) \\
 & PNI & 28,988 & 0.738 (0.726--0.749) & 0.763 (0.753--0.774) \\
\addlinespace[2pt]
MIMIC-IV sepsis & COMBIO & 28,858 & 0.781 (0.773--0.788) & --- \\
 & Best single: albumin & 28,858 & 0.758 (0.751--0.767) & --- \\
 & NLR & 28,767 /\allowbreak{} 28,858 & 0.709 (0.700--0.718) & 0.724 (0.716--0.732) \\
\end{longtable}
\tablenote{Published and refitted indices are distinct comparators. Selected index comparisons are displayed here, while the full evaluated registry, including the exploratory FIB-4 mortality comparison, is in Supplementary Table~\ref{stab5}. All five single laboratory measurements are in Supplementary Table~\ref{stab6}. Cells show AUROC (95\% percentile bootstrap CI, 1,000 resamples). The best-single identity is descriptive. Display choices do not alter model fitting or the evaluation masks. TG/HDL uses the separate HDL-complete subset, with COMBIO shown directly above. Eligible n is shown as published / refitted when those denominators differ. CI, confidence interval; FIB-4, fibrosis-4 index; APRI, aspartate aminotransferase-to-platelet ratio index; PNI, prognostic nutritional index; NLR, neutrophil-to-lymphocyte ratio; TG/HDL, triglyceride-to-high-density-lipoprotein cholesterol ratio. Index definitions and original clinical uses are in Supplementary Table~\ref{stab4}.}
\endgroup\medskip

\subsection{Learned Interpretable Comparators}\label{learned-interpretable-comparators}

COMBIO and SHAPoint had similar iron-overload AUROC point estimates (0.846 versus 0.837), whereas COMBIO was higher in ILPD (0.755 versus 0.693). In MIMIC-IV mortality, COMBIO (0.773) exceeded FasterRisk (0.754) and symbolic regression (0.737--0.752), but trailed SHAPoint (0.793) and EBM (0.824). EBM had higher point estimates than COMBIO across all three tasks, while comparisons with the other scorers varied. Full results for all compared models, including 95\% confidence intervals, are presented in Supplementary Table~\ref{stab7}.

\subsection{Exploratory Rounded-Logistic Comparison}\label{exploratory-rounded-logistic-comparison}

To assess whether simply rounding logistic-regression coefficients could produce an equally compact alternative, we compared COMBIO with a separately fitted rounded-logistic formula (Table~\ref{tab4}). Unlike the five-variable logistic benchmark, this comparator restricts the final score to two to four small integer exponents. COMBIO had higher AUROC point estimates in four of six tasks. The paired difference intervals were above zero for diabetes, MIMIC-IV mortality, and sepsis.

\FloatBarrier\par\medskip\noindent\begin{minipage}{\columnwidth}
\captionof{table}{Comparison of COMBIO and rounded logistic regression}
\label{tab4}
\centering\tablefont\setlength{\tabcolsep}{3pt}
\begin{tabular}{@{}L{\dimexpr 0.2200\linewidth-1.7600\tabcolsep\relax} L{\dimexpr 0.1100\linewidth-0.8800\tabcolsep\relax} L{\dimexpr 0.2250\linewidth-1.8000\tabcolsep\relax} L{\dimexpr 0.2250\linewidth-1.8000\tabcolsep\relax} L{\dimexpr 0.2200\linewidth-1.7600\tabcolsep\relax}@{}}
\toprule
\textbf{Task} & \textbf{Shared n} & \textbf{COMBIO AUROC (95\% CI)} & \textbf{Rounded logistic AUROC (95\% CI)} & \textbf{Difference 95\% CI} \\
\midrule
Iron overload & 1,532 & \makecell[l]{0.846\\(0.779--0.910)} & \makecell[l]{0.802\\(0.733--0.865)} & {[}-0.025, 0.122{]} \\
\addlinespace[2pt]
Five-year mortality & 1,772 & \makecell[l]{0.769\\(0.716--0.818)} & \makecell[l]{0.793\\(0.745--0.836)} & {[}-0.083, 0.030{]} \\
\addlinespace[2pt]
Undiagnosed diabetes & 1,252 & \makecell[l]{0.771\\(0.711--0.831)} & \makecell[l]{0.695\\(0.628--0.767)} & {[}0.015, 0.129{]} \\
\addlinespace[2pt]
ILPD liver disease & 175 & \makecell[l]{0.755\\(0.677--0.827)} & \makecell[l]{0.761\\(0.685--0.832)} & {[}-0.065, 0.051{]} \\
\addlinespace[2pt]
MIMIC-IV mortality & 28,988 & \makecell[l]{0.773\\(0.762--0.782)} & \makecell[l]{0.749\\(0.738--0.759)} & {[}0.015, 0.034{]} \\
\addlinespace[2pt]
MIMIC-IV sepsis & 28,830 & \makecell[l]{0.781\\(0.773--0.788)} & \makecell[l]{0.737\\(0.728--0.746)} & {[}0.035, 0.051{]} \\
\bottomrule
\end{tabular}
\tablenote{AUROC cells show 95\% percentile bootstrap intervals from 1,000 resamples on shared reserved test observations. The final column shows the paired COMBIO-minus-rounded-logistic difference interval from the same number of resamples. CI, confidence interval. Differences are COMBIO minus rounded logistic regression.}
\end{minipage}\par\medskip

\subsection{External Transportability}\label{external-transportability}

The selected formulas were applied to independent cohorts without refitting the formulas or adapting their risk mappings to target-cohort outcomes (Table~\ref{tab5}). External AUROCs were 0.777 for five-year mortality and 0.787 for the diabetes phenotype in separate NHANES cycles, and 0.698 for mortality and 0.708 for sepsis in eICU. Calibration changed: the MIMIC-derived mortality and sepsis mappings had eICU calibration slopes of 0.69 and 0.70 on their formula-complete cohorts. Internal and external estimates use different eligible populations and are not paired changes. Detailed external comparisons, reverse derivation, and hospital-level results are reported in Supplementary Results, External Transportability (Supplementary Tables~\ref{stab8}, \ref{stab9}, and \ref{stab10}).

\begin{table}[!t]
\caption{External discrimination without formula refitting}
\label{tab5}
\centering\tablefont
\setlength{\tabcolsep}{3pt}
\begin{tabular}{@{}L{\dimexpr 0.2800\linewidth-1.6800\tabcolsep\relax} L{\dimexpr 0.2500\linewidth-1.5000\tabcolsep\relax} L{\dimexpr 0.2000\linewidth-1.2000\tabcolsep\relax} L{\dimexpr 0.2700\linewidth-1.6200\tabcolsep\relax}@{}}
\toprule
\textbf{Derivation \ensuremath{\rightarrow} evaluation} & \textbf{Outcome} & \textbf{External n /\allowbreak{} events} & \textbf{AUROC (95\% CI)} \\
\midrule
NHANES 2017--18 \ensuremath{\rightarrow} 2015--16 & Undiagnosed diabetes & 4,464 /\allowbreak{} 130 & \makecell[l]{0.787\\(0.750--0.822)} \\
\addlinespace[2pt]
NHANES 2009--10 \ensuremath{\rightarrow} 2005--06 & Five-year mortality & 4,863 /\allowbreak{} 280 & \makecell[l]{0.777\\(0.743--0.807)} \\
\addlinespace[2pt]
MIMIC-IV \ensuremath{\rightarrow} eICU & In-hospital mortality & 93,697 /\allowbreak{} 9,455 & \makecell[l]{0.698\\(0.693--0.704)} \\
MIMIC-IV \ensuremath{\rightarrow} eICU & Sepsis case-finding & 97,090 /\allowbreak{} 18,255 & \makecell[l]{0.708\\(0.704--0.712)} \\
\addlinespace[2pt]
eICU \ensuremath{\rightarrow} MIMIC-IV (reverse derivation) & In-hospital mortality & 96,037 /\allowbreak{} 6,110 & \makecell[l]{0.771\\(0.765--0.778)} \\
\bottomrule
\end{tabular}
\tablenote{AUROCs have 95\% percentile bootstrap intervals from 1,000 external-row resamples. Each row has its own source, target and complete-case cohort. Rows are not directly paired. CI, confidence interval. NHANES, National Health and Nutrition Examination Survey; MIMIC-IV, Medical Information Mart for Intensive Care IV; eICU, eICU Collaborative Research Database.}
\end{table}

Additional sensitivity and subgroup analyses are reported in Supplementary Results (Supplementary Tables~\ref{stab9}, \ref{stab11}, and \ref{stab12}).

\subsection{Calibration and Decision-Curve Performance}\label{calibration-and-decision-curve-performance}

\begin{figure*}[!t]
\centering
\includegraphics[width=\textwidth]{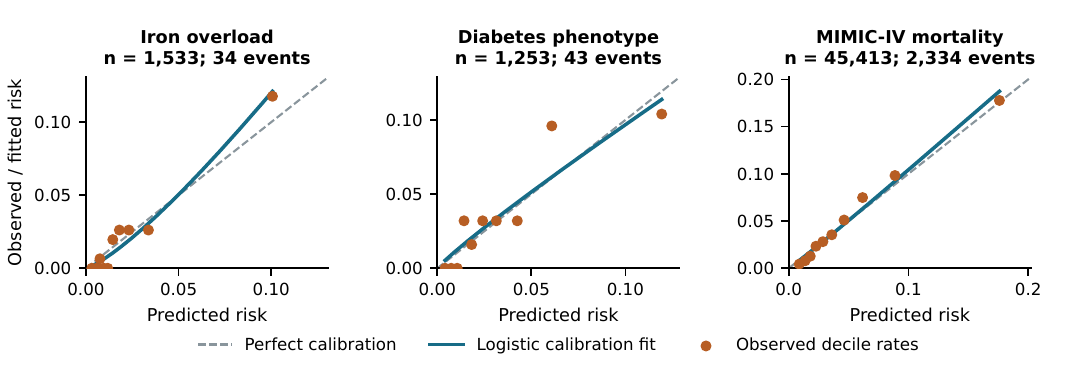}
\caption{Calibration of COMBIO on formula-complete test cohorts. Points show observed decile event rates, solid lines show fitted logistic calibration curves, and dashed lines indicate perfect calibration.}
\label{fig3}
\end{figure*}

\begin{figure*}[!t]
\centering
\includegraphics[width=\textwidth]{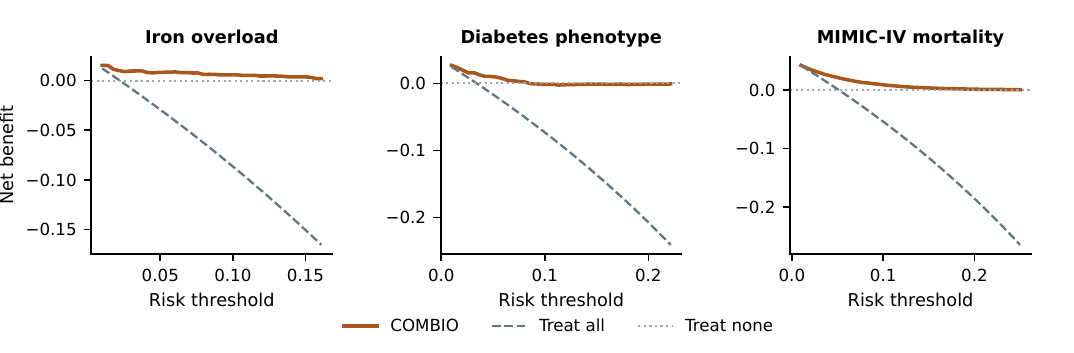}
\caption{Decision-curve net benefit for COMBIO on formula-complete test cohorts, compared with treat-all and treat-none strategies.}
\label{fig4}
\end{figure*}

AUROC describes ranking but does not show whether estimated risks are well calibrated or useful at specific decision thresholds, particularly when outcomes are uncommon. We therefore evaluated calibration, decision-curve net benefit and operating-point metrics using formulas set before evaluation and source-fitted probability mappings (Table~\ref{tab6}, Figs.~\ref{fig3}--\ref{fig4}).

\begin{table*}[!t]
\caption{Secondary performance on formula-complete test cohorts}
\label{tab6}
\centering\tablefont
\setlength{\tabcolsep}{3pt}
\begin{tabular}{@{}L{\dimexpr 0.2300\linewidth-2.3000\tabcolsep\relax} L{\dimexpr 0.1800\linewidth-1.8000\tabcolsep\relax} L{\dimexpr 0.1200\linewidth-1.2000\tabcolsep\relax} L{\dimexpr 0.1200\linewidth-1.2000\tabcolsep\relax} L{\dimexpr 0.1500\linewidth-1.5000\tabcolsep\relax} L{\dimexpr 0.2000\linewidth-2.0000\tabcolsep\relax}@{}}
\toprule
\textbf{Task} & \textbf{Test n /\allowbreak{} events} & \textbf{AUPR} & \textbf{Brier} & \textbf{Calibration Slope} & \textbf{Top-5\% PPV} \\
\midrule
Iron overload & 1,533 /\allowbreak{} 34 & 0.223 & 0.0195 & 1.26 & 0.221 \\
\addlinespace[2pt]
Five-year mortality & 1,775 /\allowbreak{} 95 & 0.234 & 0.0458 & 0.97 & 0.303 \\
\addlinespace[2pt]
Undiagnosed diabetes & 1,253 /\allowbreak{} 43 & 0.088 & 0.0326 & 0.91 & 0.063 \\
\addlinespace[2pt]
ILPD liver disease & 175 /\allowbreak{} 125 & 0.891 & 0.1727 & 1.11 & 1.000 \\
\addlinespace[2pt]
MIMIC-IV mortality & 45,413 /\allowbreak{} 2,334 & 0.161 & 0.0460 & 1.05 & 0.207 \\
\addlinespace[2pt]
MIMIC-IV sepsis & 30,577 /\allowbreak{} 4,022 & 0.343 & 0.1010 & 0.98 & 0.476 \\
\bottomrule
\end{tabular}
\tablenote{These eligible subsets differ from the primary common-case comparison subsets in Table~\ref{tab2}. AUPR, Brier score, calibration slope and top-5\% positive predictive value (PPV) are descriptive. Full capacity-rule metrics are in Supplementary Table~\ref{stab13}. AUPR, area under the precision--recall curve; PPV, positive predictive value.}
\end{table*}

Calibration slope/intercept on formula-complete test cohorts were 1.05/0.15 for MIMIC-IV mortality, 0.91/\ensuremath{-}0.23 for diabetes and 1.26/0.78 for iron overload, with Brier scores 0.046, 0.033 and 0.020 (Fig.~\ref{fig3}). On independent formula-complete eICU cohorts, slopes were 0.69 for MIMIC-derived mortality and 0.70 for sepsis. When the highest-risk 5\% were flagged, PPV was 0.207 for MIMIC-IV mortality and 0.221 for iron overload. These analyses use larger formula-complete subsets than some primary comparisons and do not establish a clinical action threshold.

Decision-curve analysis compares the mapped COMBIO score with treat-all and treat-none strategies over the displayed threshold ranges (Fig.~\ref{fig4}). COMBIO had positive net benefit over part of those ranges. Benefit approached zero at higher thresholds for iron overload and diabetes. These retrospective curves provide threshold context without establishing an intervention or optimal threshold.

Precision--recall performance and case enrichment are reported in Supplementary Results, Precision--Recall Performance and Case Enrichment (Supplementary Table~\ref{stab13}).

\FloatBarrier\section{Discussion}\label{discussion}

\subsection{Principal Findings}\label{principal-findings}

We present COMBIO, a composite biomarker that compresses information from two to four selected laboratory measurements into a formula that can be calculated directly. This represents a compactness--performance trade-off rather than a universal accuracy gain: it often had higher AUROC point estimates than single measurements and clinical indices, and remained within \ensuremath{\pm}0.03 AUROC of logistic regression using the five candidate variables.

The advantage was not universal. Glucose-containing predictors were stronger for HbA1c-defined diabetes, where glucose was available but deliberately excluded to evaluate a glucose-free alternative. Full XGBoost had higher point estimates in four tasks, and the smaller cohorts did not provide conclusive evidence of a difference from the strongest single measurement.

External evaluation further showed that risk ranking could transfer across cohorts and hospitals even when calibration changed. The high-coverage MIMIC-IV analysis retained mortality discrimination in a broader admission cohort, while the pooled-cycle diabetes result was less stable. These findings support COMBIO as a candidate for compact risk stratification. Prospective workflow evaluation, local validation and recalibration are required to determine whether the gain in simplicity justifies the loss of predictive flexibility.

\subsection{Comparison With Prior Work}\label{comparison-with-prior-work}

COMBIO occupies an intermediate position between manually derived clinical indices and more flexible learned interpretable models. Like the neutrophil-to-lymphocyte ratio, FIB-4 and TyG \cite{r1,r2,r3}, it yields a single formula calculated from a small number of measurements, but its variables and exponents are selected automatically from data. Additive point systems such as FasterRisk \cite{r10} and SHAPoint \cite{r12} assign and sum points; COMBIO applies one expression to measured values. Compared with feature-construction methods such as FEAT \cite{r19} and score-table generators such as SymScore \cite{r20}, COMBIO restricts the search to a single monomial with small integer exponents.

\subsection{Limitations}\label{limitations}

COMBIO is a constrained form of logistic regression on log-transformed laboratory values. Its contribution is a compact representation, and the benchmark does not demonstrate a universal gain in predictive accuracy.

Complete-case requirements limit applicability. Albumin and other measurements may be obtained selectively in sicker patients, so performance in eligible records may not represent the full source population. The high-coverage sensitivity analysis changed both the input panel and the evaluation population and therefore does not isolate the effect of complete-case selection. The tasks also have different outcome and cohort definitions, which limits direct comparisons between tasks.

The strongest single laboratory measurement was selected from test AUROC and is a descriptive comparator. Subgroup AUROCs likewise do not establish equal discrimination or calibration across demographic groups.

\section{Conclusions}\label{conclusions}

COMBIO provides a reproducible procedure for converting risk information in routine laboratory measurements into a compact closed-form biomarker. Across six tasks, it retained most of the discrimination of logistic regression using the five candidate variables, with an AUROC difference of no more than 0.03, and often had higher point estimates than a single laboratory measurement or selected clinical index. Clearer advantages were concentrated in the larger MIMIC-IV analyses. In the exploratory three-task learned-scorer comparison, full-panel EBM had higher AUROC point estimates in all three tasks. Rankings against SHAPoint, FasterRisk and symbolic regression varied by task. External discrimination transferred across cohorts, but calibration and hospital performance varied. Its value is a transparent formula using two to four measurements, rather than an increase in predictive accuracy. Prospective evaluation and local recalibration are required before clinical use.

\section*{Acknowledgments}\label{acknowledgments}

We thank the providers and participants of NHANES, MIMIC-IV, the eICU Collaborative Research Database, and the UCI Indian Liver Patient Dataset for making these data available. Generative artificial intelligence tools, including Codex (OpenAI), assisted with language, organization, and formatting across the manuscript and supplement, and with analysis code. The authors reviewed and edited all material prepared with these tools and take full responsibility for the analyses, scientific content, and publication.

\section*{Declarations}

\noindent\textbf{Funding.} This research was supported by The Israel Science Foundation (grant No. 2672/24). The funder had no role in the design of the study; the collection, analysis, or interpretation of data; or in writing the manuscript.

\noindent\textbf{Conflicts of interest.} None declared.

\noindent\textbf{Data and code availability.} NHANES and the UCI Indian Liver Patient Dataset are publicly available. MIMIC-IV and eICU are available to credentialed researchers through PhysioNet. The COMBIO Python package and a public-data example are available at \url{https://github.com/TomerMeirman/combio}. Analysis scripts and a fully specified benchmark will be released upon publication.

\noindent\textbf{Author contributions.} \textbf{Tomer D. Meirman:} Conceptualization, methodology, software, formal analysis, data curation, writing---original draft, visualization. \textbf{Lior S. Rokach and Bracha Shapira:} Methodology, review of the analyses, scientific advising. \textbf{Seffi Cohen:} Methodology, formal analysis, writing---review and editing. \textbf{Noa Dagan:} Conceptualization, supervision, writing---review and editing.

\FloatBarrier\bibliographystyle{IEEEtran}
\bibliography{references}
\clearpage
\section*{Supplementary Material}
\setcounter{table}{0}\renewcommand{\thetable}{S\arabic{table}}\renewcommand{\theHtable}{S\arabic{table}}
\section*{Supplementary Note 1. Reporting and Analysis Scope}

This supplement reports the current Shapley additive explanations (SHAP)-based composite monomial biomarker (COMBIO) analyses. The six outcome-labelled cohorts were drawn from the available source records; no formal sample-size calculation was performed. A test partition was reserved before fitting. Formula discovery used inner training and validation subsets of the remaining source observations; Medical Information Mart for Intensive Care IV (MIMIC-IV) partitions kept admissions from the same patient together. Formula selection used validation area under the receiver operating characteristic curve (AUROC), while the same reserved test partition was used for method comparisons. Different reported analyses have different complete-case requirements, so their denominators are reported separately. The highest observed single-laboratory AUROC and the main-table choice of clinical indices are descriptive displays chosen after evaluation. The full exploratory fibrosis-4 index (FIB-4) mortality comparison is retained in the supplement.

Scores used source-recorded numeric laboratory values, with values below 0.001 clipped before log transformation. A source-fitted logistic mapping was used for probability-based analyses; no evaluation outcomes were used to update external predictions. AUROC confidence intervals use 1,000 percentile bootstrap resamples. MIMIC-IV resampling is by admission and may understate uncertainty when patients contribute more than one admission. National Health and Nutrition Examination Survey (NHANES) estimates are unweighted benchmark results, not national survey estimates. Subgroup and post hoc sensitivity results are exploratory, without a family-wise multiplicity adjustment. Source units must be matched before applying a formula to new data.

Formula input scales in the NHANES and hospital analyses followed the source-recorded values: mean corpuscular volume (MCV) in fL; red cell distribution width (RDW) and lymphocyte differential as percentages; glucose, triglycerides, bilirubin and blood urea nitrogen (BUN) in mg/dL; gamma-glutamyl transferase (GGT), alanine aminotransferase (ALT) and aspartate aminotransferase (AST) in U/L; albumin and haemoglobin in g/dL; high-sensitivity C-reactive protein (hsCRP) in mg/L; and white blood cell count (WBC) and platelets in recorded cell-count units (10\textsuperscript{3}/\ensuremath{\mu}L). The Indian Liver Patient Dataset (ILPD) formula used the source-recorded numeric scales of alkaline phosphatase, total bilirubin, ALT and direct bilirubin; their physical units were not independently verified. Unit conversion must precede formula or risk-mapping reuse.

\section*{Supplementary Results}

\subsection*{Task-Specific Interpretation and Exploratory Comparisons}

Several task-specific choices are important for interpretation. The primary diabetes estimate derives from NHANES 2017--18, while a separate pooled 2015--16 and 2017--18 derivation had test AUROC 0.749 (n = 2,565; 84 events). The triglyceride-to-high-density-lipoprotein cholesterol ratio (TG/HDL) is evaluated on its additional HDL-complete subset. Glucose is excluded from COMBIO\textquotesingle s diabetes candidate set to assess a glucose-free predictor of the glycated hemoglobin (HbA1c)-defined phenotype. For five-year mortality, FIB-4 includes age, whereas COMBIO uses laboratory measurements alone. The sepsis outcome is International Classification of Diseases (ICD)-defined case-finding, not clinically adjudicated sepsis.

\subsection*{Exploratory FIB-4 Mortality Comparison}

FIB-4 includes age and was included as an exploratory mortality comparator because of its reported mortality association \cite{r40}. Full published and refitted comparisons are in Supplementary Table~\ref{stab5}. The separate age-eligible analysis is in Supplementary Table~\ref{stab11}.

\subsection*{External Transportability}

We assessed transportability by applying each frozen formula and its source-fitted risk mapping to independent data without external refitting (main Table~5). The diabetes formula derived in NHANES 2017--18 had test AUROC 0.771 and external AUROC 0.787 in 2015--16. For five-year mortality, external AUROC was 0.777 in NHANES 2005--06 on the common-case subset. The formula-complete cohort had AUROC 0.778 and calibration slope 0.95. MIMIC-IV mortality and sepsis formulas transferred to eICU with AUROCs of 0.698 and 0.708 on their primary external comparison subsets. Their formula-complete calibration slopes were 0.69 and 0.70. Test and external estimates use different cohorts and are not paired changes.

On the eICU comparison subsets, five-candidate logistic regression had AUROC 0.709 (0.704--0.714) for mortality and 0.708 (0.704--0.712) for sepsis. Published BUN/albumin had AUROC 0.675 (0.670--0.681) for mortality, and published neutrophil-to-lymphocyte ratio (NLR) had AUROC 0.681 (0.677--0.686) for sepsis. A MIMIC-fitted BUN/albumin model had AUROC 0.699 (0.694--0.705) for eICU mortality. These intervals use the stated eligible subsets and 1,000 admission-level bootstrap resamples (Supplementary Table~\ref{stab8}).

Sensitivity analysis: reverse derivation and external evaluation for in-hospital mortality. To assess dependence on the MIMIC-IV derivation cohort, we repeated COMBIO training in eICU and applied the frozen result to MIMIC-IV. The eICU formula was AST\textsuperscript{2} \ensuremath{\times} BUN / (albumin\textsuperscript{2} \ensuremath{\times} ALT), with eICU test AUROC 0.729. It retained BUN as a positive term and squared albumin as a negative term, as in the MIMIC-IV formula, while its other terms differed. Applied to 96,037 formula-complete MIMIC-IV admissions without refitting, it achieved AUROC 0.771. This reverse transfer complements the forward MIMIC-IV-to-eICU analysis but uses a different source cohort and evaluation subset (Supplementary Table~\ref{stab9}).

\subsubsection*{Hospital-level Variation in Transportability}

To assess variation across hospitals, we estimated AUROC separately within eICU hospitals meeting the minimum counts specified in Methods. The mortality analysis included 112 of 199 hospitals. Median hospital AUROC was 0.703 (interquartile range (IQR) 0.674--0.725), and the random-effects pooled estimate was 0.704 (95\% confidence interval (CI) 0.696--0.711, I\textsuperscript{2} 55\%). The sepsis analysis included 125 of 196 hospitals, with median AUROC 0.721 (IQR 0.693--0.745) and pooled AUROC 0.720 (95\% CI 0.713--0.727, I\textsuperscript{2} 64\%). These formula-complete hospital subsets differ from the primary external cohorts. The observed between-hospital variation supports local evaluation before clinical use (Supplementary Table~\ref{stab10}).

\subsection*{MIMIC-IV Mortality in a High-Coverage Cohort: Assessing Complete-Case Selection}

The primary MIMIC-IV mortality comparison required albumin and laboratories used by the evaluated clinical indices. Its 28,988 common-case test admissions represent about 6.7\% of the 434,542 source admissions, with mortality prevalence 6.5\% versus 2.4\% overall. Selective measurement could therefore limit generalisability. The formula-complete secondary cohort is larger, but it too may reflect informative test ordering.

To examine this concern, we repeated COMBIO training with ten laboratory measurements recorded in at least 90\% of admissions: platelets, haemoglobin, WBC, MCV, RDW, creatinine, BUN, potassium, sodium and glucose. This retained 399,658 of 434,542 admissions (92\%). The separately derived formula, BUN \ensuremath{\times} WBC \ensuremath{\times} RDW\textsuperscript{2} / platelets, achieved reserved test AUROC 0.786 (n = 119,435). Both the input panel and evaluation population changed from the primary analysis, so the AUROC difference does not isolate complete-case selection. Residual bias remains possible (Supplementary Table~\ref{stab9}).

\subsection*{Precision-Recall Performance and Case Enrichment in Low-Prevalence Outcomes}

Because several outcomes were uncommon, AUROC alone can obscure low absolute precision. Full area under the precision--recall curve (AUPR) and enrichment results are in Supplementary Table~\ref{stab13}. Across the six tasks, AUPR ranged from 1.25 to 10.04 times outcome prevalence, and the highest-risk 5\% had 1.4- to 10.0-fold enrichment. ILPD's 71\% case prevalence makes its high absolute AUPR less directly comparable with screening tasks. These estimates describe enrichment within eligible retrospective cohorts, not prospective screening yield.

\par\begingroup\tablefont\setlength{\tabcolsep}{2pt}
\begin{longtable}{@{}L{\dimexpr 0.2000\linewidth-2.0000\tabcolsep\relax} L{\dimexpr 0.1800\linewidth-1.8000\tabcolsep\relax} L{\dimexpr 0.1200\linewidth-1.2000\tabcolsep\relax} L{\dimexpr 0.1600\linewidth-1.6000\tabcolsep\relax} L{\dimexpr 0.1700\linewidth-1.7000\tabcolsep\relax} L{\dimexpr 0.1700\linewidth-1.7000\tabcolsep\relax}@{}}
\caption{Analysis-cohort flow}\label{stab1}\\
\toprule \textbf{Task} & \textbf{Labelled N/\allowbreak{}events} & \textbf{Inner train N} & \textbf{Validation N} & \textbf{Test N/\allowbreak{}events} & \textbf{Common-case test N/\allowbreak{}events} \\\midrule\endfirsthead
\multicolumn{6}{l}{\textit{Table S1 (continued)}}\\\toprule \textbf{Task} & \textbf{Labelled N/\allowbreak{}events} & \textbf{Inner train N} & \textbf{Validation N} & \textbf{Test N/\allowbreak{}events} & \textbf{Common-case test N/\allowbreak{}events} \\\midrule\endhead
\midrule\multicolumn{6}{r}{\textit{Continued on next page}}\\\endfoot
\bottomrule\endlastfoot
Iron overload & 5,122 /\newline 113 & 2,509 & 1,076 & 1,537 /\newline 34 & 1,532 /\newline 34 \\
Five-year mortality & 6,346 /\newline 363 & 3,109 & 1,333 & 1,904 /\newline 109 & 1,772 /\newline 94 \\
Undiagnosed diabetes & 4,290 /\newline 151 & 2,102 & 901 & 1,287 /\newline 45 & 1,253 /\newline 43 \\
ILPD liver disease & 583 /\newline 416 & 285 & 123 & 175 /\newline 125 & 175 /\newline 125 \\
MIMIC-IV mortality & 434,542 /\newline 10,596 & 213,049 & 90,859 & 130,634 /\newline 3,264 & 28,988 /\newline 1,891 \\
MIMIC-IV sepsis & 434,542 /\newline 22,109 & 213,049 & 90,859 & 130,634 /\newline 6,646 & 28,858 /\newline 3,898 \\
\end{longtable}
\endgroup\par\addvspace{5pt}
\noindent\footnotesize Counts begin at the outcome-labelled analysis cohort, not raw database enrolment. The reserved test N precedes complete-case filtering. Source training was split 70:30 into inner training and validation. MIMIC-IV splits were grouped by patient; the other splits were by row.\par\normalsize\medskip
\par\begingroup\tablefont\setlength{\tabcolsep}{2pt}
\begin{longtable}{@{}L{\dimexpr 0.2000\linewidth-1.2000\tabcolsep\relax} L{\dimexpr 0.1500\linewidth-0.9000\tabcolsep\relax} L{\dimexpr 0.3000\linewidth-1.8000\tabcolsep\relax} L{\dimexpr 0.3500\linewidth-2.1000\tabcolsep\relax}@{}}
\caption{Cohort age and sex}\label{stab2}\\
\toprule \textbf{Group} & \textbf{N} & \textbf{Age, years} & \textbf{Female /\allowbreak{} known sex} \\\midrule\endfirsthead
\multicolumn{4}{l}{\textit{Table S2 (continued)}}\\\toprule \textbf{Group} & \textbf{N} & \textbf{Age, years} & \textbf{Female /\allowbreak{} known sex} \\\midrule\endhead
\midrule\multicolumn{4}{r}{\textit{Continued on next page}}\\\endfoot
\bottomrule\endlastfoot
\multicolumn{4}{l}{\textbf{Iron overload}}\\*
Overall & 5,122 & 52 (34--65) & 2,665 /\allowbreak{} 5,122 (52.0\%) \\
No outcome & 5,009 & 52 (34--65) & 2,635 /\allowbreak{} 5,009 (52.6\%) \\
Outcome & 113 & 58 (45--69) & 30 /\allowbreak{} 113 (26.6\%) \\
\multicolumn{4}{l}{\textbf{Five-year mortality}}\\*
Overall & 6,346 & 47 (32--63) & 3,259 /\allowbreak{} 6,346 (51.4\%) \\
No outcome & 5,983 & 46 (31--61) & 3,120 /\allowbreak{} 5,983 (52.1\%) \\
Outcome & 363 & 75 (64--80) & 139 /\allowbreak{} 363 (38.3\%) \\
\multicolumn{4}{l}{\textbf{Undiagnosed diabetes}}\\*
Overall & 4,290 & 47 (31--62) & 2,285 /\allowbreak{} 4,290 (53.3\%) \\
No outcome & 4,139 & 46 (31--62) & 2,194 /\allowbreak{} 4,139 (53.0\%) \\
Outcome & 151 & 58 (50--67) & 91 /\allowbreak{} 151 (60.3\%) \\
\multicolumn{4}{l}{\textbf{ILPD liver disease}}\\*
Overall & 583 & 45 (33--58) & 142 /\allowbreak{} 583 (24.4\%) \\
No outcome & 167 & 40 (28--55) & 50 /\allowbreak{} 167 (29.9\%) \\
Outcome & 416 & 46 (34--58) & 92 /\allowbreak{} 416 (22.1\%) \\
\multicolumn{4}{l}{\textbf{MIMIC-IV mortality}}\\*
Overall & 434,542 & 61 (47--73) & 223,059 /\allowbreak{} 434,542 (51.3\%) \\
No outcome & 423,946 & 61 (47--73) & 218,252 /\allowbreak{} 423,946 (51.5\%) \\
Outcome & 10,596 & 71 (59--81) & 4,807 /\allowbreak{} 10,596 (45.4\%) \\
\multicolumn{4}{l}{\textbf{MIMIC-IV sepsis}}\\*
Overall & 434,542 & 61 (47--73) & 223,059 /\allowbreak{} 434,542 (51.3\%) \\
No outcome & 412,433 & 61 (47--73) & 213,069 /\allowbreak{} 412,433 (51.7\%) \\
Outcome & 22,109 & 65 (53--76) & 9,990 /\allowbreak{} 22,109 (45.2\%) \\
\end{longtable}
\endgroup\par\addvspace{5pt}
\noindent\footnotesize Characteristics refer to full outcome-labelled cohorts. Age is median (interquartile range) among those with recorded age. Female percentages use the known-sex denominator. MIMIC-IV anchor\_age is de-identified and top-coded. In ILPD, the source-coded sex variable was interpreted as male when its string began with M and female otherwise.\par\normalsize\medskip
\par\begingroup\tablefont\setlength{\tabcolsep}{2pt}
\begin{longtable}{@{}L{\dimexpr 0.0900\linewidth-0.7200\tabcolsep\relax} L{\dimexpr 0.3300\linewidth-2.6400\tabcolsep\relax} L{\dimexpr 0.1800\linewidth-1.4400\tabcolsep\relax} L{\dimexpr 0.2500\linewidth-2.0000\tabcolsep\relax} L{\dimexpr 0.1500\linewidth-1.2000\tabcolsep\relax}@{}}
\caption{SHAP-ranked candidates and exponents}\label{stab3}\\
\toprule \textbf{Rank} & \textbf{Measurement} & \textbf{Mean \textbar SHAP\textbar{}} & \textbf{Trend} & \textbf{Exponent} \\\midrule\endfirsthead
\multicolumn{5}{l}{\textit{Table S3 (continued)}}\\\toprule \textbf{Rank} & \textbf{Measurement} & \textbf{Mean \textbar SHAP\textbar{}} & \textbf{Trend} & \textbf{Exponent} \\\midrule\endhead
\midrule\multicolumn{5}{r}{\textit{Continued on next page}}\\\endfoot
\bottomrule\endlastfoot
\multicolumn{5}{l}{\textbf{Iron overload}}\\*
1 & GGT & 0.563 & positive & 1 \\
2 & bilirubin & 0.419 & positive & 1 \\
3 & MCV & 0.407 & ambiguous & 2 \\
4 & ALT & 0.319 & ambiguous & 0 \\
5 & platelets & 0.273 & ambiguous & 0 \\
\multicolumn{5}{l}{\textbf{Five-year mortality}}\\*
1 & BUN & 0.564 & ambiguous & 1 \\
2 & RDW & 0.513 & positive & 2 \\
3 & platelets & 0.409 & ambiguous & -1 \\
4 & ALT & 0.323 & negative & 0 \\
5 & glucose & 0.294 & ambiguous & 1 \\
\multicolumn{5}{l}{\textbf{Undiagnosed diabetes}}\\*
1 & TG & 0.705 & ambiguous & 2 \\
2 & GGT & 0.610 & ambiguous & 1 \\
3 & albumin & 0.402 & ambiguous & -2 \\
4 & creatinine & 0.370 & ambiguous & 0 \\
5 & hsCRP & 0.298 & ambiguous & 1 \\
\multicolumn{5}{l}{\textbf{ILPD liver disease}}\\*
1 & alkaline phosphatase & 0.784 & ambiguous & 2 \\
2 & total protein & 0.686 & ambiguous & 0 \\
3 & total bilirubin & 0.533 & ambiguous & 2 \\
4 & direct bilirubin & 0.473 & ambiguous & -1 \\
5 & ALT & 0.461 & ambiguous & 2 \\
\multicolumn{5}{l}{\textbf{MIMIC-IV mortality}}\\*
1 & BUN & 0.412 & positive & 1 \\
2 & albumin & 0.324 & negative & -2 \\
3 & WBC & 0.309 & positive & 0 \\
4 & glucose & 0.256 & positive & 1 \\
5 & RDW & 0.250 & positive & 2 \\
\multicolumn{5}{l}{\textbf{MIMIC-IV sepsis}}\\*
1 & WBC & 0.418 & positive & 0 \\
2 & lymphocyte \% & 0.380 & negative & -1 \\
3 & albumin & 0.284 & negative & -2 \\
4 & bilirubin & 0.217 & positive & 0 \\
5 & haemoglobin & 0.211 & negative & -1 \\
\end{longtable}
\endgroup\par\addvspace{5pt}
\noindent\footnotesize The five candidates in each task were ranked by mean absolute Tree SHAP value on inner validation. The directional trend describes the association between a measurement and its SHAP contribution to positive-class log odds; it is not a causal effect. A zero selected exponent omits the measurement from the final formula.\par\normalsize\medskip
\par\begingroup\tablefont\setlength{\tabcolsep}{2pt}
\begin{longtable}{@{}L{\dimexpr 0.1700\linewidth-0.6800\tabcolsep\relax} L{\dimexpr 0.3300\linewidth-1.3200\tabcolsep\relax} L{\dimexpr 0.5000\linewidth-2.0000\tabcolsep\relax}@{}}
\caption{Clinical-index definitions and task relevance}\label{stab4}\\
\toprule \textbf{Index} & \textbf{Definition} & \textbf{Original use and scope} \\\midrule\endfirsthead
\multicolumn{3}{l}{\textit{Table S4 (continued)}}\\\toprule \textbf{Index} & \textbf{Definition} & \textbf{Original use and scope} \\\midrule\endhead
\midrule\multicolumn{3}{r}{\textit{Continued on next page}}\\\endfoot
\bottomrule\endlastfoot
PNI & 10 \ensuremath{\times} albumin (g/\allowbreak{}dL) + 0.005 \ensuremath{\times} absolute lymphocytes (cells/\allowbreak{}\ensuremath{\mu}L) & Prognostic nutritional index \cite{r5}. Mortality evidence in community and critically ill cohorts \cite{r38,r39}. Lower values generally indicate poorer nutritional and immune status. \\
FIB-4 & \(\dfrac{\mathrm{Age}\times\mathrm{AST}}{\mathrm{platelets}\times\sqrt{\mathrm{ALT}}}\) & Liver-fibrosis assessment \cite{r2}. Its mortality association \cite{r40} motivates an exploratory mortality comparison. Its iron-overload application is exploratory. \\
APRI & \(\dfrac{100\,(\mathrm{AST}/\allowbreak{}40)}{\mathrm{platelets}}\) & AST-to-platelet ratio index for liver-fibrosis assessment \cite{r4}. The implementation uses 40 U/\allowbreak{}L as the AST upper reference limit. Iron-overload use is exploratory. \\
De Ritis ratio & AST /\allowbreak{} ALT & Originally described in viral hepatitis \cite{r33}. Applications to broad ILPD classification and iron overload are exploratory. \\
NLR & Neutrophils /\allowbreak{} lymphocytes & Neutrophil-to-lymphocyte ratio. Inflammation context \cite{r1} and sepsis-discrimination evidence \cite{r36}. \\
TG/\allowbreak{}HDL & Triglycerides /\allowbreak{} HDL cholesterol & Insulin-resistance marker \cite{r37}. This differs from the HbA1c-defined diabetes phenotype. Both measurements use compatible units. \\
BUN/\allowbreak{}albumin ratio & Blood urea nitrogen /\allowbreak{} serum albumin & Mortality prediction in community-acquired pneumonia \cite{r35}. Application to other mortality cohorts is exploratory. \\
CAR & C-reactive protein /\allowbreak{} serum albumin & C-reactive protein-to-albumin ratio. Acute medical admission outcome context \cite{r34}. The community five-year-mortality application is exploratory. \\
Bilirubin/\allowbreak{}albumin ratio & Total bilirubin /\allowbreak{} serum albumin & Exploratory comparison for broad ILPD classification. \\
\end{longtable}
\endgroup\par\addvspace{5pt}
\noindent\footnotesize Formula scales follow the recorded implementation. Age is in years, AST and ALT are in U/L, and platelets are in 10\textsuperscript{9}/L for FIB-4 and APRI. Absolute lymphocyte counts are derived from WBC and lymphocyte percentage where necessary. Refitted-index models use fitted coefficients on log-transformed components and are distinct from the original scores or clinical cutoffs.\par\normalsize\medskip
\par\begingroup\tablefont\setlength{\tabcolsep}{2pt}
\begin{longtable}{@{}L{\dimexpr 0.2800\linewidth-1.6800\tabcolsep\relax} L{\dimexpr 0.1700\linewidth-1.0200\tabcolsep\relax} L{\dimexpr 0.2750\linewidth-1.6500\tabcolsep\relax} L{\dimexpr 0.2750\linewidth-1.6500\tabcolsep\relax}@{}}
\caption{All clinical-index comparisons}\label{stab5}\\
\toprule \textbf{Candidate/\allowbreak{}index} & \textbf{Eligible n} & \textbf{Measured/\allowbreak{}published AUROC (95\% CI)} & \textbf{Refitted AUROC (95\% CI)} \\\midrule\endfirsthead
\multicolumn{4}{l}{\textit{Table S5 (continued)}}\\\toprule \textbf{Candidate/\allowbreak{}index} & \textbf{Eligible n} & \textbf{Measured/\allowbreak{}published AUROC (95\% CI)} & \textbf{Refitted AUROC (95\% CI)} \\\midrule\endhead
\midrule\multicolumn{4}{r}{\textit{Continued on next page}}\\\endfoot
\bottomrule\endlastfoot
\multicolumn{4}{l}{\textbf{Iron overload}}\\*
COMBIO & 1,532 & 0.846 (0.779--0.910) & --- \\
FIB-4 & 1,532 & 0.708 (0.604--0.801) & 0.786 (0.700--0.857) \\
APRI & 1,532 & 0.767 (0.669--0.850) & 0.769 (0.671--0.855) \\
De Ritis ratio & 1,532 & 0.582 (0.476--0.684) & 0.761 (0.665--0.853) \\
\multicolumn{4}{l}{\textbf{Five-year mortality}}\\*
COMBIO & 1,772 & 0.769 (0.716--0.818) & --- \\
FIB-4 & 1,772 & 0.834 (0.793--0.871) & 0.866 (0.827--0.903) \\
PNI & 1,772 & 0.720 (0.658--0.775) & 0.762 (0.707--0.812) \\
BUN/\allowbreak{}albumin ratio & 1,772 & 0.721 (0.663--0.779) & 0.738 (0.679--0.794) \\
NLR & 1,772 & 0.682 (0.621--0.737) & 0.719 (0.660--0.773) \\
CAR & 1,772 & 0.631 (0.569--0.689) & 0.674 (0.614--0.732) \\
\multicolumn{4}{l}{\textbf{Undiagnosed diabetes}}\\*
COMBIO & 1,253 & 0.771 (0.710--0.825) & --- \\
COMBIO (HDL-complete) & 1,252 & 0.771 (0.711--0.831) & --- \\
TG/\allowbreak{}HDL (HDL-complete) & 1,252 & 0.675 (0.613--0.738) & 0.675 (0.612--0.739) \\
\multicolumn{4}{l}{\textbf{ILPD liver disease}}\\*
COMBIO & 175 & 0.755 (0.677--0.827) & --- \\
Bilirubin/\allowbreak{}albumin ratio & 175 & 0.774 (0.703--0.841) & 0.741 (0.667--0.813) \\
De Ritis ratio & 175 & 0.657 (0.569--0.741) & 0.739 (0.657--0.816) \\
\multicolumn{4}{l}{\textbf{MIMIC-IV mortality}}\\*
COMBIO & 28,988 & 0.773 (0.762--0.782) & --- \\
BUN/\allowbreak{}albumin ratio & 28,988 & 0.746 (0.735--0.757) & 0.779 (0.769--0.789) \\
PNI & 28,988 & 0.738 (0.726--0.749) & 0.763 (0.753--0.774) \\
FIB-4 & 28,988 & 0.700 (0.688--0.712) & 0.731 (0.719--0.742) \\
NLR & 28,895 /\allowbreak{} 28,988 & 0.687 (0.673--0.699) & 0.709 (0.697--0.721) \\
\multicolumn{4}{l}{\textbf{MIMIC-IV sepsis}}\\*
COMBIO & 28,858 & 0.781 (0.773--0.788) & --- \\
NLR & 28,767 /\allowbreak{} 28,858 & 0.709 (0.700--0.718) & 0.724 (0.716--0.732) \\
\end{longtable}
\endgroup\par\addvspace{5pt}
\noindent\footnotesize AUROC (95\% percentile bootstrap confidence interval) from 1,000 resamples on the stated eligible test subset. Published and refitted indices are distinct comparators. Eligible n is shown separately when denominators differ. The FIB-4 mortality comparison is exploratory.\par\normalsize\medskip
\par\begingroup\tablefont\setlength{\tabcolsep}{2pt}
\begin{longtable}{@{}L{\dimexpr 0.3400\linewidth-1.3600\tabcolsep\relax} L{\dimexpr 0.1800\linewidth-0.7200\tabcolsep\relax} L{\dimexpr 0.4800\linewidth-1.9200\tabcolsep\relax}@{}}
\caption{All five selected single measurements}\label{stab6}\\
\toprule \textbf{Measurement} & \textbf{Eligible n} & \textbf{AUROC (95\% CI)} \\\midrule\endfirsthead
\multicolumn{3}{l}{\textit{Table S6 (continued)}}\\\toprule \textbf{Measurement} & \textbf{Eligible n} & \textbf{AUROC (95\% CI)} \\\midrule\endhead
\midrule\multicolumn{3}{r}{\textit{Continued on next page}}\\\endfoot
\bottomrule\endlastfoot
\multicolumn{3}{l}{\textbf{Iron overload}}\\*
GGT & 1,532 & 0.766 (0.675--0.858) \\
bilirubin & 1,532 & 0.781 (0.710--0.852) \\
MCV & 1,532 & 0.695 (0.591--0.793) \\
ALT & 1,532 & 0.752 (0.657--0.845) \\
platelets & 1,532 & 0.657 (0.557--0.752) \\
\multicolumn{3}{l}{\textbf{Five-year mortality}}\\*
BUN & 1,772 & 0.703 (0.641--0.764) \\
RDW & 1,772 & 0.741 (0.689--0.787) \\
platelets & 1,772 & 0.612 (0.545--0.678) \\
ALT & 1,772 & 0.595 (0.519--0.661) \\
glucose & 1,772 & 0.683 (0.621--0.743) \\
\multicolumn{3}{l}{\textbf{Undiagnosed diabetes}}\\*
TG & 1,253 & 0.669 (0.603--0.737) \\
GGT & 1,253 & 0.690 (0.624--0.757) \\
albumin & 1,253 & 0.652 (0.560--0.738) \\
creatinine & 1,253 & 0.507 (0.407--0.609) \\
hsCRP & 1,253 & 0.725 (0.652--0.786) \\
\multicolumn{3}{l}{\textbf{ILPD liver disease}}\\*
alkaline phosphatase & 175 & 0.664 (0.577--0.749) \\
total protein & 175 & 0.451 (0.358--0.536) \\
total bilirubin & 175 & 0.735 (0.660--0.807) \\
direct bilirubin & 175 & 0.732 (0.661--0.802) \\
ALT & 175 & 0.711 (0.633--0.787) \\
\multicolumn{3}{l}{\textbf{MIMIC-IV mortality}}\\*
BUN & 28,988 & 0.695 (0.683--0.707) \\
albumin & 28,988 & 0.737 (0.726--0.747) \\
WBC & 28,988 & 0.623 (0.608--0.636) \\
glucose & 28,988 & 0.567 (0.552--0.581) \\
RDW & 28,988 & 0.647 (0.634--0.659) \\
\multicolumn{3}{l}{\textbf{MIMIC-IV sepsis}}\\*
WBC & 28,858 & 0.652 (0.642--0.663) \\
lymphocyte \% & 28,858 & 0.719 (0.710--0.727) \\
albumin & 28,858 & 0.758 (0.751--0.767) \\
bilirubin & 28,858 & 0.605 (0.596--0.615) \\
haemoglobin & 28,858 & 0.605 (0.595--0.614) \\
\end{longtable}
\endgroup\par\addvspace{5pt}
\noindent\footnotesize AUROCs (95\% percentile bootstrap confidence intervals, 1,000 resamples) use the common test subset. The five candidates were selected by validation Tree SHAP magnitude, with single-measurement risk orientation set in inner training. The main article reports the highest observed test AUROC descriptively.\par\normalsize\medskip
\par\begingroup\tablefont\setlength{\tabcolsep}{2pt}
\begin{longtable}{@{}L{\dimexpr 0.3200\linewidth-1.9200\tabcolsep\relax} L{\dimexpr 0.1300\linewidth-0.7800\tabcolsep\relax} L{\dimexpr 0.2500\linewidth-1.5000\tabcolsep\relax} L{\dimexpr 0.3000\linewidth-1.8000\tabcolsep\relax}@{}}
\caption{Learned-scorer comparison}\label{stab7}\\
\toprule \textbf{Score} & \textbf{Fitting N} & \textbf{Test N/\allowbreak{}events} & \textbf{AUROC (95\% CI)} \\\midrule\endfirsthead
\multicolumn{4}{l}{\textit{Table S7 (continued)}}\\\toprule \textbf{Score} & \textbf{Fitting N} & \textbf{Test N/\allowbreak{}events} & \textbf{AUROC (95\% CI)} \\\midrule\endhead
\midrule\multicolumn{4}{r}{\textit{Continued on next page}}\\\endfoot
\bottomrule\endlastfoot
\multicolumn{4}{l}{\textbf{Iron overload}}\\*
COMBIO & --- & 1,532 /\allowbreak{} 34 & 0.846 (0.779--0.910) \\
Symbolic regression, retained inputs & 3,585 & 1,532 /\allowbreak{} 34 & 0.832 (0.764--0.895) \\
Symbolic regression, full panel & 3,585 & 1,532 /\allowbreak{} 34 & 0.856 (0.797--0.907) \\
SHAPoint, full panel & 3,585 & 1,532 /\allowbreak{} 34 & 0.837 (0.769--0.898) \\
FasterRisk, full panel & 3,585 & 1,532 /\allowbreak{} 34 & 0.824 (0.752--0.884) \\
Additive EBM, full panel & 3,585 & 1,532 /\allowbreak{} 34 & 0.856 (0.788--0.909) \\
\multicolumn{4}{l}{\textbf{ILPD liver disease}}\\*
COMBIO & --- & 175 /\allowbreak{} 125 & 0.755 (0.677--0.827) \\
Symbolic regression, retained inputs & 408 & 175 /\allowbreak{} 125 & 0.776 (0.705--0.842) \\
Symbolic regression, full panel & 408 & 175 /\allowbreak{} 125 & 0.792 (0.731--0.853) \\
SHAPoint, full panel & 408 & 175 /\allowbreak{} 125 & 0.693 (0.603--0.772) \\
FasterRisk, full panel & 408 & 175 /\allowbreak{} 125 & 0.785 (0.717--0.850) \\
Additive EBM, full panel & 408 & 175 /\allowbreak{} 125 & 0.782 (0.708--0.846) \\
\multicolumn{4}{l}{\textbf{MIMIC-IV mortality}}\\*
COMBIO & --- & 28,988 /\allowbreak{} 1,891 & 0.773 (0.762--0.782) \\
Symbolic regression, retained inputs & 20,000 & 28,988 /\allowbreak{} 1,891 & 0.737 (0.726--0.747) \\
Symbolic regression, full panel & 20,000 & 28,988 /\allowbreak{} 1,891 & 0.752 (0.742--0.761) \\
SHAPoint, full panel & 30,000 & 28,988 /\allowbreak{} 1,891 & 0.793 (0.784--0.802) \\
FasterRisk, full panel & 60,000 & 28,988 /\allowbreak{} 1,891 & 0.754 (0.744--0.764) \\
Additive EBM, full panel & 60,000 & 28,988 /\allowbreak{} 1,891 & 0.824 (0.816--0.832) \\
\end{longtable}
\endgroup\par\addvspace{5pt}
\noindent\footnotesize This exploratory comparison uses each task\textquotesingle s primary common-case test subset for every row. Models were fitted on source training data, with the MIMIC-IV fitting caps described in Methods. Symbolic regression used either the retained COMBIO measurements or the full laboratory panel. Entries are AUROC (95\% percentile bootstrap CI; 1,000 participant or admission resamples). Intervals describe each score separately and do not establish paired superiority. EBM, explainable boosting machine.\par\normalsize\medskip
\par\begingroup\tablefont\setlength{\tabcolsep}{2pt}
\begin{longtable}{@{}L{\dimexpr 0.2000\linewidth-1.2000\tabcolsep\relax} L{\dimexpr 0.3000\linewidth-1.8000\tabcolsep\relax} L{\dimexpr 0.2000\linewidth-1.2000\tabcolsep\relax} L{\dimexpr 0.3000\linewidth-1.8000\tabcolsep\relax}@{}}
\caption{External eICU comparator discrimination}\label{stab8}\\
\toprule \textbf{Outcome} & \textbf{Comparator} & \textbf{eICU n/\allowbreak{}events} & \textbf{AUROC (95\% CI)} \\\midrule\endfirsthead
\multicolumn{4}{l}{\textit{Table S8 (continued)}}\\\toprule \textbf{Outcome} & \textbf{Comparator} & \textbf{eICU n/\allowbreak{}events} & \textbf{AUROC (95\% CI)} \\\midrule\endhead
\midrule\multicolumn{4}{r}{\textit{Continued on next page}}\\\endfoot
\bottomrule\endlastfoot
Mortality & Five-measurement logistic & 93,697 /\allowbreak{} 9,455 & 0.709 (0.704--0.714) \\
Mortality & Published BUN/\allowbreak{}albumin & 93,697 /\allowbreak{} 9,455 & 0.675 (0.670--0.681) \\
Mortality & Refitted BUN/\allowbreak{}albumin & 93,697 /\allowbreak{} 9,455 & 0.699 (0.694--0.705) \\
Sepsis & Five-measurement logistic & 97,090 /\allowbreak{} 18,255 & 0.708 (0.704--0.712) \\
Sepsis & Published NLR & 96,938 /\allowbreak{} 18,196 & 0.681 (0.677--0.686) \\
\end{longtable}
\endgroup\par\addvspace{5pt}
\noindent\footnotesize The published clinical-index scores, source-fitted clinical-index model, and five-measurement logistic models were evaluated on their stated eICU complete-case subsets. Entries are AUROC (95\% percentile bootstrap CI) from 1,000 admission resamples.\par\normalsize\medskip
\par\begingroup\tablefont\setlength{\tabcolsep}{2pt}
\begin{longtable}{@{}L{\dimexpr 0.2700\linewidth-1.6200\tabcolsep\relax} L{\dimexpr 0.4000\linewidth-2.4000\tabcolsep\relax} L{\dimexpr 0.2000\linewidth-1.2000\tabcolsep\relax} L{\dimexpr 0.1300\linewidth-0.7800\tabcolsep\relax}@{}}
\caption{Separately derived robustness formulas}\label{stab9}\\
\toprule \textbf{Analysis} & \textbf{Formula} & \textbf{Evaluation N/\allowbreak{}events} & \textbf{AUROC} \\\midrule\endfirsthead
\multicolumn{4}{l}{\textit{Table S9 (continued)}}\\\toprule \textbf{Analysis} & \textbf{Formula} & \textbf{Evaluation N/\allowbreak{}events} & \textbf{AUROC} \\\midrule\endhead
\midrule\multicolumn{4}{r}{\textit{Continued on next page}}\\\endfoot
\bottomrule\endlastfoot
MIMIC-IV mortality, high coverage & BUN \ensuremath{\times} WBC \ensuremath{\times} RDW\textsuperscript{2} /\allowbreak{} platelets & 119,435 /\allowbreak{} 3,130 & 0.786 \\
NHANES diabetes, pooled cycles & TG\textsuperscript{2} \ensuremath{\times} GGT\textsuperscript{2} \ensuremath{\times} hsCRP /\allowbreak{} AST & 2,565 /\allowbreak{} 84 & 0.749 \\
eICU mortality \ensuremath{\rightarrow} MIMIC-IV & AST\textsuperscript{2} \ensuremath{\times} BUN /\allowbreak{} (albumin\textsuperscript{2} \ensuremath{\times} ALT) & 96,037 /\allowbreak{} 6,110 & 0.771 \\
\end{longtable}
\endgroup\par\addvspace{5pt}
\noindent\footnotesize These searches used their stated source cohorts and candidate panels. Their test estimates are not paired changes from the six primary analyses. Reverse transfer applies the eICU-derived formula to MIMIC-IV without target-cohort refitting.\par\normalsize\medskip
\par\begingroup\tablefont\setlength{\tabcolsep}{2pt}
\begin{longtable}{@{}L{\dimexpr 0.3600\linewidth-1.4400\tabcolsep\relax} L{\dimexpr 0.3200\linewidth-1.2800\tabcolsep\relax} L{\dimexpr 0.3200\linewidth-1.2800\tabcolsep\relax}@{}}
\caption{Hospital-level external variation}\label{stab10}\\
\toprule Statistic & MIMIC-IV mortality & MIMIC-IV sepsis \\\midrule\endfirsthead
\multicolumn{3}{l}{\textit{Table S10 (continued)}}\\\toprule Statistic & MIMIC-IV mortality & MIMIC-IV sepsis \\\midrule\endhead
\midrule\multicolumn{3}{r}{\textit{Continued on next page}}\\\endfoot
\bottomrule\endlastfoot
\textbf{Hospitals} & 112 /\allowbreak{} 199 & 125 /\allowbreak{} 196 \\
\textbf{Stays/\allowbreak{}events} & 115,468 /\allowbreak{} 12,143 & 98,795 /\allowbreak{} 19,599 \\
\textbf{Median (IQR)} & 0.703 (0.674--0.725) & 0.721 (0.693--0.745) \\
\textbf{Pooled AUROC (95\% CI)} & 0.704 (0.696--0.711) & 0.720 (0.713--0.727) \\
\textbf{I\textsuperscript{2}} & 55\% & 64\% \\
\textbf{\ensuremath{\tau}} & 0.027 & 0.030 \\
\end{longtable}
\endgroup\par\addvspace{5pt}
\noindent\footnotesize eICU hospitals qualified with at least 100 stays, 25 events and 25 non-events. The pooled AUROC and interval come from a DerSimonian--Laird random-effects analysis \cite{r44} on the AUROC scale. I\textsuperscript{2} and \ensuremath{\tau} summarise between-hospital heterogeneity \cite{r45}.\par\normalsize\medskip
\par\begingroup\tablefont\setlength{\tabcolsep}{2pt}
\begin{longtable}{@{}L{\dimexpr 0.4500\linewidth-1.8000\tabcolsep\relax} L{\dimexpr 0.2000\linewidth-0.8000\tabcolsep\relax} L{\dimexpr 0.3500\linewidth-1.4000\tabcolsep\relax}@{}}
\caption{Exploratory age-eligible mortality formula}\label{stab11}\\
\toprule \textbf{Score} & \textbf{Test n /\allowbreak{} deaths} & \textbf{AUROC (95\% CI)} \\\midrule\endfirsthead
\multicolumn{3}{l}{\textit{Table S11 (continued)}}\\\toprule \textbf{Score} & \textbf{Test n /\allowbreak{} deaths} & \textbf{AUROC (95\% CI)} \\\midrule\endhead
\midrule\multicolumn{3}{r}{\textit{Continued on next page}}\\\endfoot
\bottomrule\endlastfoot
Laboratory-only COMBIO & 1,772 /\allowbreak{} 94 & 0.769 (0.715--0.821) \\
Age-eligible COMBIO & 1,772 /\allowbreak{} 94 & 0.865 (0.831--0.896) \\
Published FIB-4 & 1,772 /\allowbreak{} 94 & 0.834 (0.796--0.870) \\
Refitted FIB-4 & 1,772 /\allowbreak{} 94 & 0.866 (0.828--0.901) \\
\end{longtable}
\endgroup\par\addvspace{5pt}
\noindent\footnotesize This exploratory sensitivity analysis was specified after the primary comparison had been reviewed. The original inner training, validation and reserved test partition was reused. Age was added to the candidate panel and ranked first by mean absolute Tree SHAP. The selected formula was age\textsuperscript{2} \ensuremath{\times} RDW\textsuperscript{2} \ensuremath{\times} WBC / platelets. All estimates use the same 1,772 test participants, including 94 deaths. This analysis does not replace the laboratory-only primary method.\par\normalsize\medskip
Paired age-eligible COMBIO minus refitted FIB-4 AUROC difference: -0.002 (95\% percentile bootstrap CI -0.024 to +0.021; 1,000 resamples). The post hoc timing limits confirmatory interpretation.\par\vspace{6pt}
\par\begingroup\tablefont\setlength{\tabcolsep}{2pt}
\begin{longtable}{@{}L{\dimexpr 0.3400\linewidth-2.7200\tabcolsep\relax} L{\dimexpr 0.1900\linewidth-1.5200\tabcolsep\relax} L{\dimexpr 0.1300\linewidth-1.0400\tabcolsep\relax} L{\dimexpr 0.1700\linewidth-1.3600\tabcolsep\relax} L{\dimexpr 0.1700\linewidth-1.3600\tabcolsep\relax}@{}}
\caption{Exploratory MIMIC-IV subgroups}\label{stab12}\\
\toprule \textbf{Recorded group} & \textbf{N} & \textbf{Events} & \textbf{AUROC} & \textbf{Brier} \\\midrule\endfirsthead
\multicolumn{5}{l}{\textit{Table S12 (continued)}}\\\toprule \textbf{Recorded group} & \textbf{N} & \textbf{Events} & \textbf{AUROC} & \textbf{Brier} \\\midrule\endhead
\midrule\multicolumn{5}{r}{\textit{Continued on next page}}\\\endfoot
\bottomrule\endlastfoot
\multicolumn{5}{l}{\textbf{MIMIC-IV mortality}}\\*
Overall & 45,413 & 2,334 & 0.771 & 0.046 \\
Female & 21,274 & 1,041 & 0.780 & 0.044 \\
Male & 24,139 & 1,293 & 0.763 & 0.048 \\
Age \textless65 & 25,907 & 924 & 0.778 & 0.033 \\
Age \ensuremath{\geq}65 & 19,506 & 1,410 & 0.745 & 0.063 \\
White & 30,805 & 1,451 & 0.782 & 0.042 \\
Black & 6,163 & 233 & 0.745 & 0.036 \\
Hispanic & 2,371 & 83 & 0.771 & 0.033 \\
Asian & 1,619 & 75 & 0.739 & 0.043 \\
Other/\allowbreak{}unknown & 4,455 & 492 & 0.763 & 0.094 \\
\multicolumn{5}{l}{\textbf{MIMIC-IV sepsis}}\\*
Overall & 30,577 & 4,022 & 0.780 & 0.101 \\
Female & 14,465 & 1,808 & 0.786 & 0.096 \\
Male & 16,112 & 2,214 & 0.774 & 0.105 \\
Age \textless65 & 17,808 & 2,017 & 0.781 & 0.090 \\
Age \ensuremath{\geq}65 & 12,769 & 2,005 & 0.772 & 0.116 \\
White & 20,491 & 2,648 & 0.775 & 0.101 \\
Black & 4,191 & 470 & 0.774 & 0.089 \\
Hispanic & 1,641 & 181 & 0.827 & 0.085 \\
Asian & 1,119 & 161 & 0.779 & 0.107 \\
Other/\allowbreak{}unknown & 3,135 & 562 & 0.783 & 0.126 \\
\end{longtable}
\endgroup\par\addvspace{5pt}
\noindent\footnotesize Formula-complete test admissions are grouped by recorded sex, age and race. These descriptive results do not test equality or establish fairness. Brier scores use source-fitted probability mappings.\par\normalsize\medskip
\par\begingroup\tablefont\setlength{\tabcolsep}{2pt}
\begin{longtable}{@{}L{\dimexpr 0.2000\linewidth-2.4000\tabcolsep\relax} L{\dimexpr 0.1500\linewidth-1.8000\tabcolsep\relax} L{\dimexpr 0.1000\linewidth-1.2000\tabcolsep\relax} L{\dimexpr 0.1000\linewidth-1.2000\tabcolsep\relax} L{\dimexpr 0.1100\linewidth-1.3200\tabcolsep\relax} L{\dimexpr 0.1700\linewidth-2.0400\tabcolsep\relax} L{\dimexpr 0.1700\linewidth-2.0400\tabcolsep\relax}@{}}
\caption{Secondary test performance}\label{stab13}\\
\toprule \textbf{Task} & \textbf{Test N/\allowbreak{}events} & \textbf{AUPR} & \textbf{Brier} & \textbf{Cal. slope} & \textbf{Top 5\% PPV /\allowbreak{} sens.} & \textbf{Top 10\% PPV /\allowbreak{} sens.} \\\midrule\endfirsthead
\multicolumn{7}{l}{\textit{Table S13 (continued)}}\\\toprule \textbf{Task} & \textbf{Test N/\allowbreak{}events} & \textbf{AUPR} & \textbf{Brier} & \textbf{Cal. slope} & \textbf{Top 5\% PPV /\allowbreak{} sens.} & \textbf{Top 10\% PPV /\allowbreak{} sens.} \\\midrule\endhead
\midrule\multicolumn{7}{r}{\textit{Continued on next page}}\\\endfoot
\bottomrule\endlastfoot
Iron overload & 1,533 /\allowbreak{} 34 & 0.223 & 0.020 & 1.26 & 0.221 /\allowbreak{} 0.500 & 0.117 /\allowbreak{} 0.529 \\
Five-year mortality & 1,775 /\allowbreak{} 95 & 0.234 & 0.046 & 0.97 & 0.303 /\allowbreak{} 0.284 & 0.236 /\allowbreak{} 0.442 \\
Undiagnosed diabetes & 1,253 /\allowbreak{} 43 & 0.088 & 0.033 & 0.91 & 0.063 /\allowbreak{} 0.093 & 0.103 /\allowbreak{} 0.302 \\
ILPD liver disease & 175 /\allowbreak{} 125 & 0.891 & 0.173 & 1.11 & 1.000 /\allowbreak{} 0.072 & 1.000 /\allowbreak{} 0.144 \\
MIMIC-IV mortality & 45,413 /\allowbreak{} 2,334 & 0.161 & 0.046 & 1.05 & 0.207 /\allowbreak{} 0.201 & 0.178 /\allowbreak{} 0.346 \\
MIMIC-IV sepsis & 30,577 /\allowbreak{} 4,022 & 0.343 & 0.101 & 0.98 & 0.476 /\allowbreak{} 0.181 & 0.414 /\allowbreak{} 0.315 \\
\end{longtable}
\endgroup\par\addvspace{5pt}
\noindent\footnotesize Formula-complete test subsets differ from the primary common-case comparison subsets. AUPR is the area under the precision--recall curve. The top-5\% and top-10\% rules select fixed proportions by score rank; they are descriptive capacity rules rather than clinical thresholds. Cal., calibration; PPV, positive predictive value; sens., sensitivity.\par\normalsize\medskip

\end{document}